%% file: main.tex
\documentclass[runningheads]{llncs}

\usepackage{eccv}

\PassOptionsToPackage{pagebackref,breaklinks,colorlinks,citecolor=eccvblue}{hyperref}
\input{preamble}

\usepackage{hyperref}

\begin{document}

\title{VLM2GeoVec: Toward Universal Multimodal Embeddings for Remote Sensing}

\titlerunning{Toward Universal Multimodal Embeddings for Remote Sensing}

\author{
Emanuel Sanchez Aimar$^{1}$\orcidlink{0000-0001-9874-737X},
Gulnaz Zhambulova$^{1}$\orcidlink{0009-0008-0665-0393},\\
Fahad Shahbaz Khan$^{1,}$$^{2}$\orcidlink{0000-0002-2492-9872},
Yonghao Xu$^{1}$\orcidlink{0000-0002-6857-0152}, Michael Felsberg$^{1}$\orcidlink{0000-0002-6096-3648}\\
}

\authorrunning{E.~Sanchez Aimar et al.}

\institute{
  $^{1}$Linköping University \hspace*{5mm} $^{2}$Mohamed bin Zayed University of AI \\
\
\{name.surname\}@liu.se
}


\maketitle

\begin{abstract}
\input{abstract}
\end{abstract}

\section{Introduction}
\input{introduction}

\section{Related Work}
\input{related_work}

\section{RSMEB: A Multimodal Embedding Benchmark for Remote Sensing}
\input{benchmark}

\section{VLM2GeoVec: A Multimodal Embedder for Remote Sensing}
\input{method}

\section{Experiments}
\input{experiments}

\section{Conclusion}
\input{conclusion}

\section*{Acknowledgements}
\input{acknowledgements}

%
%
\bibliographystyle{splncs04}
\bibliography{main}


\newpage

\begin{center}
\textbf{\Large Supplementary Material}
\end{center}

\appendix
\renewcommand{\theHsection}{appendix.\Alph{section}}
\renewcommand{\theHsubsection}{appendix.\Alph{section}.\arabic{subsection}}
\input{appendix}


\end{document}

%% file: preamble.tex
\usepackage[utf8]{inputenc} 
\usepackage[T1]{fontenc}    
\usepackage{url}            
\usepackage{booktabs}       
\usepackage{amsfonts}       
\usepackage{nicefrac}       
\usepackage{graphicx}       
\usepackage{multirow}
\usepackage[table]{xcolor} 
\usepackage{pdfpages}               
\usepackage{pifont}
\usepackage{graphicx}

\usepackage{amsmath,amssymb}
\usepackage{color}
\usepackage{url}
\usepackage{wrapfig}
\usepackage{caption}

\usepackage{eccvabbrv}

\usepackage[accsupp]{axessibility}  

\usepackage{orcidlink}

\newcommand{\cmark}{\ding{51}} 
\newcommand{\xmark}{\ding{55}} 

\newcommand{\comment}[1]{}

\definecolor{LightYellow}{RGB}{255,255,204}
\definecolor{Gray}{gray}{0.9}
\definecolor{LightCyan}{rgb}{0.88,1,1}
\definecolor{Lavender}{HTML}{E6E6FA}

%% file: abstract.tex
Satellite imagery differs from natural images in viewpoint, resolution, scale variation, and the prevalence of small objects — demanding both region-level spatial reasoning and holistic scene understanding. Existing remote-sensing approaches are fragmented: dual-encoder retrieval models scale well but cannot interleave modalities, whereas generative assistants support grounding, yet are inefficient for retrieval. Benchmarks mirror this split: interleaved evaluations mainly target generative assistants, while cross-modal retrieval benchmarks target dual encoders. To bridge this gap, we introduce \textbf{RSMEB}, a unified remote sensing benchmark that evaluates cross-modal and interleaved retrieval across 21 tasks under a single ranking protocol, enabling comprehensive comparison of retrieval models on region- and geo-aware capabilities as well as conventional retrieval. As a strong reference baseline, we present \textbf{VLM2GeoVec}, an instruction-conditioned, single-encoder interleaving formulation tailored to remote sensing that packs image, text, bounding-box, and geo-coordinate tokens into one sequence and learns a unified embedding via contrastive training. Across RSMEB, VLM2GeoVec achieves $\textbf{26.6\%}$ P@1 in region-caption retrieval ($\textbf{+25}$ percentage points), $\textbf{32.5\%}$ in referring-expression retrieval ($\textbf{+19}$), and $\textbf{17.8\%}$ in semantic geo-aware retrieval ($\textbf{>3}$$\times$ prior best), while remaining competitive in conventional scene classification and text--image retrieval in zero-shot settings. Together, the proposed suite and reference baseline standardize evaluation and deliver a unified embedder for scalable retrieval and region-/geo-aware grounding. The code, the model checkpoints, and the data are available at \url{https://github.com/emasa/VLM2GeoVec}.

%% file: introduction.tex
Satellite and aerial imagery has advanced to sub‑meter resolution, capturing intricate details of urban layouts, agricultural fields, and natural landscapes in vast regions. However, this wealth of data presents unique challenges. Scenes vary dramatically in scale and viewpoint, and analyses often depend on precise spatial metadata—e.g., bounding boxes and geo-coordinates—to address tasks ranging from object detection to environmental monitoring \cite{Xia2017AID,Helber2019EuroSAT,Xia2018DOTA,Li2020DIOR,zhou2024geoground,vivanco2023geoclip}. For example, an urban planner might select a city region to locate outdoor activity spaces to ensure fair access to recreational facilities. An environmental analyst might ask to highlight the largest ship in an image to assess its impact on nearby marine habitats. A preservation planner might input the coordinates of a historic district to find architecturally significant buildings with specific roof styles. To meet these diverse needs for the handling of spatial and geospatial data along with images and language, we propose \textbf{VLM2GeoVec}, a unified multimodal retriever for Remote Sensing (RS).

\begin{wrapfigure}{r}{0.48\linewidth}
  \vspace{-20pt}
  \centering
  \includegraphics[width=\linewidth]{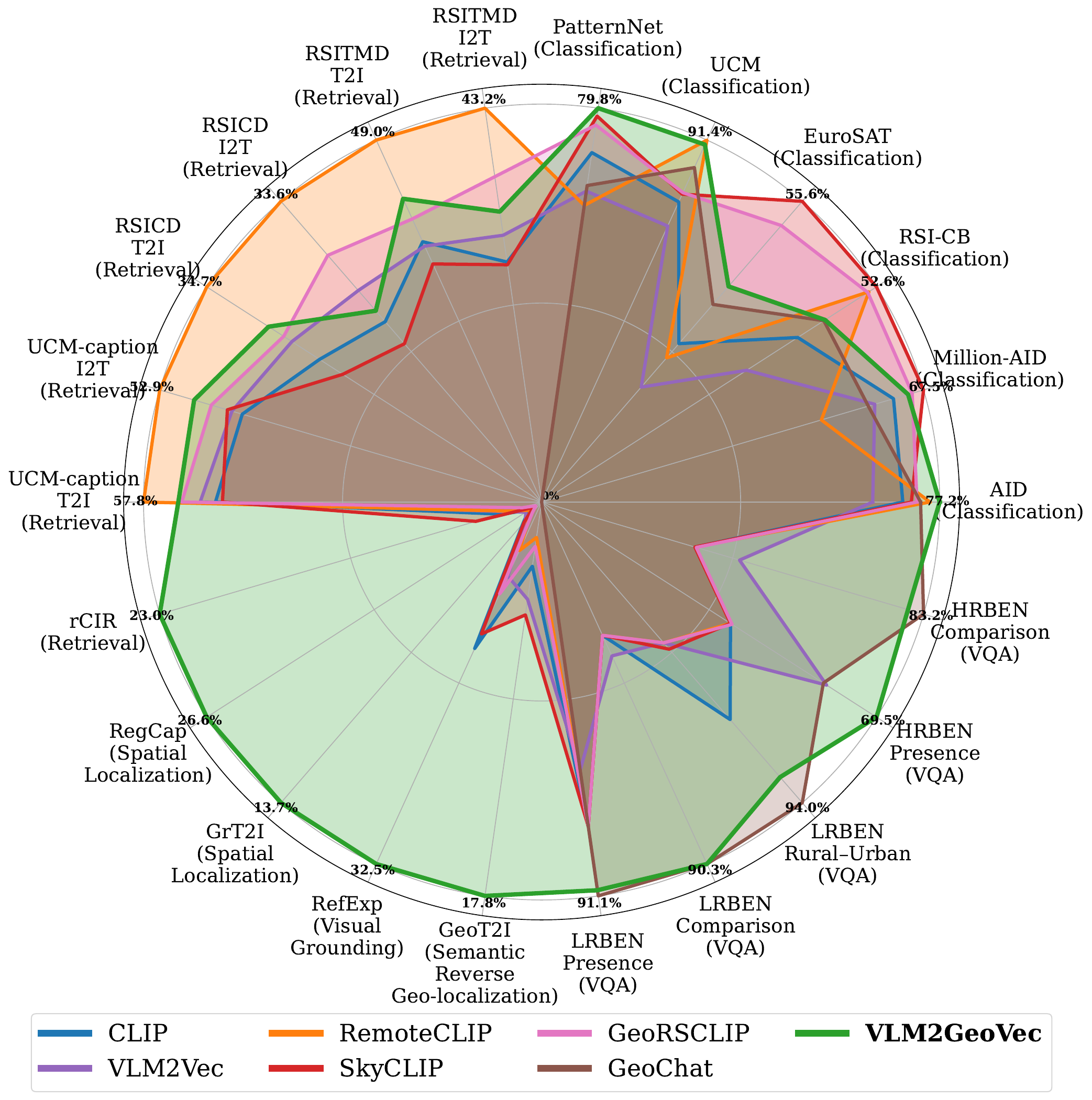}
  \caption{On the new \textbf{RSMEB} benchmark, \textbf{VLM2GeoVec} outperforms RS VLMs by wide margins on multimodal tasks and remains competitive on conventional cross-modal tasks.}
  \label{fig:rsmeb_results_radar_summary_top}
  \vspace{-20pt}
\end{wrapfigure}

Dual encoders align images and text via \emph{separate} encoders and a global similarity score, i.e., late fusion \cite{radford2021clip,jia2021align,Zhai2023SigLIP}. This setup makes it difficult for \emph{bounding-box} and \emph{geo-coordinates} tokens to influence the joint score in a single query without extra modules or multiple forward passes, so dual-encoder baselines (e.g., CLIP) underperform on region-/geo-aware retrieval (see Fig.~\ref{fig:rsmeb_results_radar_summary_top}). RS generative assistants can follow instructions and ground text \cite{Kuckreja2024GeoChat,Irvin2024TEOChat}, but decoding-based ranking is inefficient, and most do not expose an embedding head capable of large-scale semantic search. Finally, general-purpose embedders suffer from a domain shift in RS imagery, leading to weaker cross-modal retrieval~\cite{liu2024remoteclip}, which is exacerbated for interleaved multimodal tasks, as evidenced in Fig.~\ref{fig:rsmeb_results_radar_summary_top}. Table~\ref{tab:capability_matrix} contrasts various methodologies regarding their retrieval capabilities.

\begin{wraptable}{r}{0.48\linewidth}
  \vspace{-20pt}
  \centering
  \caption{\textbf{Comparative Overview of Retrieval Capabilities across Method Families.} \emph{Legend:} \textbf{Scalable} = single-embedding, scalable look-up gallery; \textbf{Interleaved} = mix image, text and metadata; \textbf{RS-aware} = trained/adapted for RS imagery.}
  \resizebox{\linewidth}{!}{%
  \begin{tabular}{lccc}
  \toprule
  \textbf{Method} & \textbf{Scalable} & \textbf{Interleaved} & \textbf{RS-aware} \\
  \midrule
  Dual Encoders (e.g.~CLIP)~\cite{radford2021clip,Zhai2023SigLIP} & \cmark & \xmark & \xmark \\
  Universal Embeddings~\cite{Jiang2024VLM2Vec,lin2024mmembed} & \cmark & \cmark & \xmark \\
  RS CLIP-like~\cite{liu2024remoteclip,wang2024skyscript,zhang2024georsclip} & \cmark & \xmark & \cmark \\
  RS Chat Assistants~\cite{Kuckreja2024GeoChat,Irvin2024TEOChat,zhan2025skyeyegpt} & \xmark & \cmark & \cmark \\
  \textbf{VLM2GeoVec (ours)} & \cmark & \cmark & \cmark \\
  \bottomrule
  \end{tabular}
  }
  \label{tab:capability_matrix}
  \vspace{-20pt}
\end{wraptable}

\begin{figure*}[t]
\centering
\includegraphics[width=\linewidth]{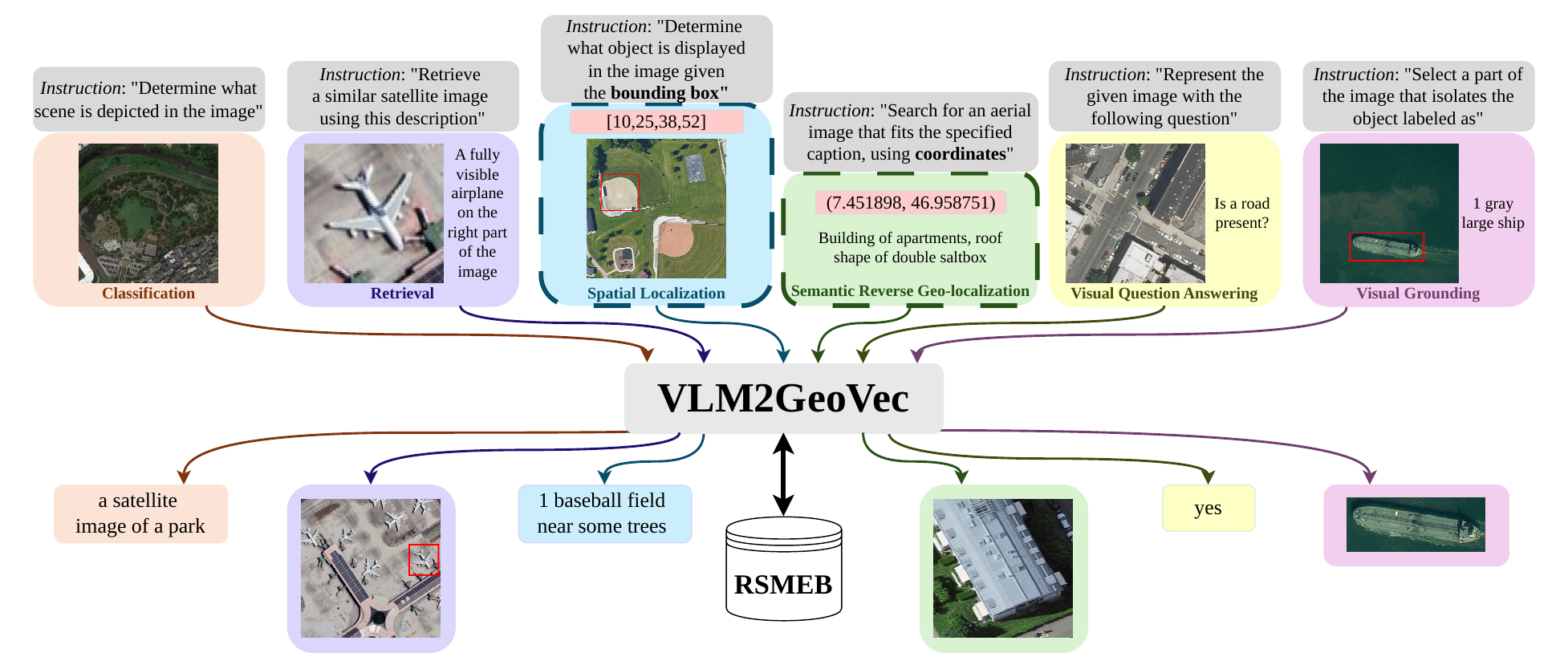}
\caption{\textbf{VLM2GeoVec} jointly embeds RGB imagery, text, bounding boxes, geo-coordinates, and a task instruction using a single multimodal encoder trained contrastively. \textbf{RSMEB} evaluates these embeddings in a unified ranking framework spanning six meta-tasks: classification, retrieval, spatial localization, semantic reverse geo-localization, visual question answering, and visual grounding.}
\label{fig:vlm2geovec_framework}
\end{figure*}

Notably, RS benchmarks remain also fragmented: scene classification~\cite{Xia2017AID,Helber2019EuroSAT} and cross-modal retrieval~\cite{Yuan2022RSITMD,Lu2017RSICD} have been the focus of embedding benchmarks, while visual question answering (VQA)~\cite{Lobry2020RSVQA} and visual grounding~\cite{Kuckreja2024GeoChat} are typically evaluated only in the context of generative assistants, excluding a holistic, retrieval-centric assessment. Universal embedding benchmarks~\cite{Muennighoff2022MTEB,Jiang2024VLM2Vec} provide cohesive multitask suites that span classification, retrieval, and grounding, but are customized for natural images and do not assess the integration of RS-specific metadata. The community lacks a unified benchmark that spans both conventional retrieval and spatial/geo-spatial reasoning. To address this fragmentation, we introduce \textbf{RSMEB}, a unified ranking benchmark that encompasses classification; cross‑modal and composed retrieval; VQA; visual grounding, spatial localization — including search for regions delimited by bounding boxes and grounded text-to-image retrieval — and semantic reverse geo-localization retrieval. Fig.~\ref{fig:vlm2geovec_framework} illustrates the new test suite.

The limitations mentioned above, highlighted by RSMEB, motivate the development of embeddings that interleave spatial and geospatial cues with images and language. To meet this need, we present \textbf{VLM2GeoVec}, an instruction-conditioned single encoder that jointly embeds imagery, text, bounding boxes, and geo-coordinates through end‑to‑end contrastive learning. Fig.~\ref{fig:vlm2geovec_framework} shows our multimodal retriever. In zero-shot evaluation, VLM2GeoVec performs on par with specialized RS dual-encoders for conventional classification and cross-modal retrieval. When queries interleave modalities, it delivers substantially larger gains (Fig.~\ref{fig:rsmeb_results_radar_summary_top}).

\noindent\textbf{Contributions.} We introduce (i) \textbf{RSMEB} — a unified, retrieval-centric suite that consolidates \textbf{21} tasks into \textbf{six} meta-tasks (classification; cross-modal and compositional retrieval; VQA; visual grounding; spatial localization; semantic reverse geo-localization) under a common ranking protocol; (ii) \textbf{VLM2GeoVec} — an \emph{instruction-conditioned single encoder} that \emph{interleaves} images, text, \emph{bounding boxes}, and \emph{geo-coordinates} into a unified contrastive embedding, enabling scalable region- and geo-aware semantic search tailored to remote-sensing applications; and (iii) a \textbf{comprehensive study} showing that \emph{general-purpose embedders fail on interleaved queries}, while our simple specialized retriever remains competitive on conventional cross-modal retrieval, with large gains on interleaved meta-tasks (e.g. VQA, visual grounding, spatial localization).

%% file: related_work.tex
\textbf{Vision-Language Models (VLMs).}
VLMs allow machines to understand both text and images~\cite{li2025benchmark}. Pioneering work like CLIP~\cite{radford2021clip} uses a dual-encoder and contrastive learning to align images and text in a shared embedding space, enabling strong zero-shot performance across vision tasks~\cite{radford2021clip,jia2021align,Zhai2023SigLIP}, later extended with the encoder-decoder architecture and generative objectives~\cite{li2022blip,yu2022coca}.

Recent works extend VLMs toward universal multimodal retrieval. For example, UniIR \cite{wei2024uniir}, MM‐Embed \cite{lin2024mmembed}, E5-V \cite{jiang2024e5v}, and VLM2Vec \cite{Jiang2024VLM2Vec} use instruction tuning to align modalities, with the latter three leveraging Multimodal Large Language Models (MLLMs) \cite{li2024llava,liu2024llavanext,li2024llavanext-strong,abdin2024phi}. Built on pretrained LLMs, MLLMs integrate vision encoders via alignment layers to exploit LLM reasoning~\cite{alayrac2022flamingo,achiam2023gpt,liu2023visual,wang2024qwen2}.
Although these models perform well on general semantic tasks, they suffer from a domain gap when transferred to RS, thus motivating the development of specialized VLMs for RS.

\textbf{Vision-Language Models for Remote Sensing.} 
One class of RS VLMs adapts the CLIP paradigm to satellite and aerial data~\cite{mo2023sclip,li2023rsclip,liu2024remoteclip,wang2024skyscript,zhang2024georsclip}. 
They are fine‐tuned on geospatial image–text pairs to improve retrieval and classification performance in RS applications. 
Some methods also contrastively incorporate RS-specific modality, such as geo-coordinates for geo-localization \cite{vivanco2023geoclip,jia2024g3,klemmer2025satclip}.

Another class of RS VLMs adopts MLLMs as generative assistants~\cite{bazi2024rsllava,luo2024skysensegpt,wang2024ringmogpt,pang2025vhm,hu2025rsgpt,zhan2025skyeyegpt}. These models handle general tasks like captioning and classification, as well as RS-specific tasks. For example, TEOChat~\cite{Irvin2024TEOChat} enables temporal reasoning; EarthGPT~\cite{zhang2024earthgpt} supports multi-sensor input; SkyEyeGPT~\cite{zhan2025skyeyegpt} and GeoChat \cite{Kuckreja2024GeoChat} offer region-level reasoning. Still, RS VLMs struggle to combine fine-grained spatial understanding with scalable retrieval over large datasets.

\textbf{Vision-Language Benchmarks.}
Multimodal benchmarks such as M-BEIR \cite{wei2024uniir} and MMEB~\cite{Jiang2024VLM2Vec} treat tasks as ranking problems, allowing modality-agnostic evaluation, but lack RS data, limiting domain relevance.
In contrast, traditional RS benchmarks are task-specific: some focus on classification~\cite{Xia2017AID,Helber2019EuroSAT}, others on retrieval~\cite{Yuan2022RSITMD,Lu2017RSICD}, VQA~\cite{Lobry2020RSVQA}, or object detection~\cite{Xia2018DOTA,Li2020DIOR}.
To address this, large-scale instruction-based RS datasets---like GeoChat~\cite{Kuckreja2024GeoChat}, TEOChatlas~\cite{Irvin2024TEOChat}, and FIT-RS~\cite{luo2024skysensegpt}---support unified multimodal reasoning. They extend general tasks with RS-specific challenges such as region-level reasoning, temporal analysis, multi-sensor fusion, and fine-grained spatial understanding.

%% file: benchmark.tex
\label{sec:rsmeb}
\subsection{Dataset Overview}

We introduce the \textbf{Remote Sensing Multimodal Embedding Benchmark }(RSMEB), a comprehensive multimodal evaluation suite that consolidates a wide range of RS datasets into a single ranking-based evaluation framework to assess both conventional and advanced multimodal capabilities. RSMEB consists of \textbf{21 tasks} organized in \textbf{6 meta-tasks}: classification, multimodal retrieval, visual question answering, visual grounding, spatial localization, and semantic reverse geo-localization (see Fig.~\ref{fig:vlm2geovec_framework}). Table~\ref{tab:dataset_overview} summarizes the associated datasets of each metatask, the input-output modalities, and the number of test-time queries and target candidates. \textbf{Well-established tasks preserve the original candidate pools}: class labels for classification, multiple ground-truth candidates for cross-modal retrieval, and predefined choices for VQA, ensuring alignment with existing evaluation protocols.

\begin{table*}[t]
\centering
\small
\caption{RSMEB meta-tasks, associated datasets, task input-output, test-time query, and target counts.}
\label{tab:dataset_overview}
\resizebox{\linewidth}{!}{
\begin{tabular}{lllrr}
\toprule
\textbf{Meta-task (\#tasks)} & \textbf{Dataset}       & \textbf{Input $\rightarrow$ Output}           & \textbf{\#Queries} & \textbf{\#Targets} \\
\midrule
\multirow{6}{*}{Classification (6)} 
    & AID \cite{Xia2017AID}                      & Image $\rightarrow$ Class label               & 2,000   & 30   \\
    & Million-AID \cite{long2021millionaid}      & Image $\rightarrow$ Class label               & 10,000  & 51   \\
    & RSI-CB \cite{li2020rsicb}                  & Image $\rightarrow$ Class label               & 24,747  & 35   \\
    & EuroSAT \cite{Helber2019EuroSAT}           & Image $\rightarrow$ Class label               & 2,700   & 10   \\
    & UCM \cite{Yang2010UCM}           & Image $\rightarrow$ Class label               & 2,100   & 21   \\
    & PatternNet \cite{zhou2018patternnet}       & Image $\rightarrow$ Class label               & 30,400  & 38   \\
\midrule
\multirow{5}{*}{Retrieval (7)}
    & RSITMD \cite{Yuan2022RSITMD}               & Image $\leftrightarrow$ Text                  & varies  & varies \\
    & RSICD \cite{Lu2017RSICD}                   & Image $\leftrightarrow$ Text                  & varies  & varies \\
    & UCM-caption \cite{qu2016ducmcaption}       & Image $\leftrightarrow$ Text                  & varies  & varies \\
    & rCIR              & Image Region + Text $\rightarrow$ Image        & 1,818 & 1115 \\
\midrule
\multirow{5}{*}{Visual Question Answering (5)}
    & LRBEN presence \cite{Lobry2020RSVQA}    & Image + Question $\rightarrow$ Answer         & 2,955   & 2     \\
    & LRBEN comparison \cite{Lobry2020RSVQA}  & Image + Question $\rightarrow$ Answer         & 4,002   & 2     \\
    & LRBEN rural/urban \cite{Lobry2020RSVQA} & Image + Question $\rightarrow$ Answer       & 100     & 2     \\
    & HRBEN presence \cite{Lobry2020RSVQA}    & Image + Question $\rightarrow$ Answer         & 58,545  & 2     \\
    & HRBEN comparison \cite{Lobry2020RSVQA}  & Image + Question $\rightarrow$ Answer         & 72,923  & 2     \\
\midrule
\multirow{1}{*}{Visual Grounding (1)}
    & RefExp    & Image + Text $\rightarrow$ Image Region  & 2,000   & 2,000 \\
\midrule
\multirow{2}{*}{Spatial Localization (2)}
    & RegCap    & Image + BBox $\rightarrow$ Text             & 2,654   & 777 \\
    & GrT2I     & Text with BBoxes $\rightarrow$ Image            & 1,622   & 1,323 \\    
\midrule
Semantic Reverse Geo-localization (1)
    & GeoT2I      & Text + (Lat,Lon) $\rightarrow$ Image       & 2,000   & 2,000 \\
\bottomrule
\end{tabular}
}

\end{table*}

\subsection{Meta‑task Design}
RSMEB tasks are formulated as ranking problems in which a retriever receives a query with one or more inputs (e.g.,~image, text, bounding box, or geo‑coordinates),
paired with an (optional) natural‑language task instruction, and must retrieve the correct set of targets from a pool of candidates.  

For \textbf{classification}, a single image is matched to its class label, with candidates equal to the set of class names. Similar to previous work~\cite{radford2021clip,liu2024remoteclip,wang2024skyscript}, we utilize an ensemble of 20 label prompts (e.g.,~\textit{a satellite image of }\texttt{[class name]}). 

The \textbf{retrieval} meta-task contains cross-modal and composed retrieval tasks. \textit{Cross-modal retrieval} alternates between image-to-text and text-to-image matching, using the original distractor pool of each dataset. Each query may have multiple positive matches (e.g., five captions in RSICD~\cite{Lu2017RSICD}). Our new~\textit{region‑based Composed Image Retrieval} (rCIR) task, sourced from~\cite{luo2024skysensegpt}, which combines a cropped region with a free-text modifier (e.g.,~\textit{with thicker smoke plume}) to retrieve full target images, permitting one correct response.

In \textbf{visual question answering}, we evaluate on the RSVQA-LRBEN and RSVQA-HRBEN~\cite{Lobry2020RSVQA} datasets, covering multiple-choice questions. Each example provides a satellite image and a question, and the model must select the correct answer from the fixed candidate set for that dataset. 

\textbf{Visual grounding} includes the \textit{referring-expression retrieval}, inspired by~\cite{Kuckreja2024GeoChat,Jiang2024VLM2Vec}, in which a full-scene image is given along with a description of the region, and the corresponding region must be identified from an image collection.  

In \textbf{spatial localization}, we introduce two tasks—\textit{region-caption retrieval} (RegCap) and \textit{grounded text-to-image retrieval} (GrT2I), sourced from~\cite{Kuckreja2024GeoChat} and~\cite{luo2024skysensegpt}, respectively—where RegCap requires retrieving a region's caption given an image and a bounding box, and GrT2I requires selecting the correct full-scene image given a caption annotated with bounding-box coordinates.  

Finally, within \textbf{semantic reverse geo‑localization}, the new~\textit{geo-localized text-to-image retrieval} (GeoT2I), sourced from~\cite{wang2024skyscript}, pairs latitude-longitude with semantic information (e.g., "Find a satellite image near \texttt{(34.052275, 118.243739)} showing a baseball stadium.") to select the correct image from geographically relevant candidates.

These tests assess the model’s ability to integrate spatial and geographical cues with images and language for accurate grounding and retrieval. 
Appendix~\ref{sec:evaluation_data_extra} details benchmark construction, filtering, and provenance. Having introduced \textbf{RSMEB}, we next present \textbf{VLM2GeoVec} as a reference model designed to support its heterogeneous retrieval interfaces within a single embedding architecture.

%% file: method.tex
In this section, we describe the unified architecture, our instruction‑conditioned contrastive learning framework, and the construction of a multimodal RS pretraining corpus.

\subsection{Unified Encoder with Multimodal Interleaving}

\begin{figure}[t]
  \centering
    \includegraphics[width=0.8\linewidth,trim={20 20 20 20}]{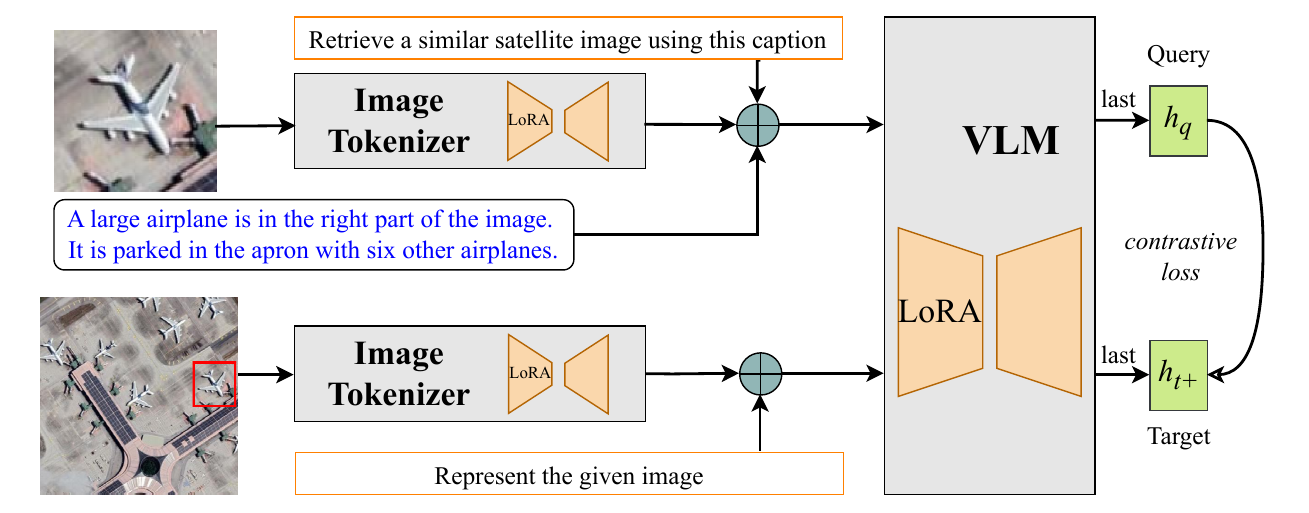}
  \caption{\textbf{Overview of VLM2GeoVec}, showing interleaved inputs (including visual tokens and text tokens, but optionally bounding boxes and geo-coordinates) encoded by a frozen VLM backbone with LoRA adapters and trained end-to-end via a contrastive objective.}
\label{fig:vlm2geovec_architecture}
\end{figure}

RS demands both precise spatial reasoning ("where") and rich semantic understanding ("what") across a diverse set of applications. To tackle these challenges with a single model, we propose \textbf{VLM2GeoVec}, an instruction-conditioned multimodal embedder for remote sensing, that can process a unified stream of interleaved tokens—images, text, bounding boxes, and geo-coordinates—and learns a joint representation via contrastive learning, as illustrated in Fig.~\ref{fig:vlm2geovec_architecture}.

Inspired by the success of instruction-following vision-language generative models~\cite{liu2024llavanext,wang2024qwen2}, we adopt a VLM as our backbone. To achieve efficient end-to-end fine-tuning, we inject lightweight low-rank adaptation (LoRA~\cite{hu2021lora}) into self-attention and MLP layers. Furthermore, we observe that general-purpose universal embeddings (e.g., VLM2Vec~\cite{Jiang2024VLM2Vec}) serve as a good initialization for our RS embedder. Hence, we integrate pre-trained LoRA adapters with the base VLM before training, and initialize a new set of LoRA weights for domain adaptation. We ablate this design decision in Sec.~\ref{sec:discussion_ablations}. To simplify training, the input images are resampled to a uniform resolution of $336\times336$~\cite{Jiang2024VLM2Vec}, subdivided into $14\times14$ patches, and tokenized with a ViT-L-14 backbone~\cite{wang2024qwen2}.

VLM2GeoVec interleaves up to four modalities in a single token stream: \textbf{visual tokens}, \textbf{descriptive textual tokens} (e.g.,~captions), axis-aligned normalized \textbf{bounding boxes} ($[x_{min},y_{min},x_{max},y_{max}]$) with integer coordinates in $[0,100]$~\cite{Irvin2024TEOChat}, and \textbf{latitude-longitude coordinates} encoded as float tuples (e.g.,~\texttt{(34.052275, 118.243739)}. No special tokens or metadata embeddings are used. A task instruction describes the search intent in natural language~\cite{lin2024mmembed,Jiang2024VLM2Vec}. Finally, we derive the embedding for each sequence from its final token for contrastive alignment~\cite{Jiang2024VLM2Vec}.

\textbf{Why interleaving helps.} Dual encoders process image, text, and spatial/geo cues separately and only merge them via a global similarity score, so boxes and coordinates have limited effect. Instead, VLM2GeoVec encodes RGB, text, boxes, and coordinates as a single token stream, letting spatial and geo tokens directly shape the embedding via self-attention. Strong gains in RefExp, RegCap, GrT2I, and GeoT2I (Table~\ref{tab:grounding_evaluation}) demonstrate the effectiveness of token-level integration for region- and geo-aware retrieval.

\subsection{Instruction-conditioned Contrastive Learning} 
Following~\cite{wei2024uniir,lin2024mmembed,Jiang2024VLM2Vec}, we adopt the InfoNCE loss~\cite{oord2018infonce_contrastive} calculated over in-batch negatives. Given a batch of \(N\) paired sequences \((q_i, t_i^+)\), we pass them through the VLM encoder to produce query and target embeddings \(h_{q_i}\) and \(h_{t_i^+}\). Then our contrastive loss is defined as 
\begin{equation}
\mathcal{L} = -\sum_{i} \log \frac{\exp\bigl(\cos(h_{q_i},h_{t_i^+})/\tau\bigr)}{\exp\bigl(\cos(h_{q_i},h_{t_i^+})/\tau\bigr) + \sum_{t^-\in\mathcal{N}} \exp\bigl(\cos(h_{q_i},h_{t^-})/\tau\bigr)},
\label{eq:vlm2geovec_loss}
\end{equation}
where $\mathcal{N}$ denotes the set of all negatives, $\cos(\cdot,\cdot)$ denotes the cosine similarity, and \(\tau\) is the temperature hyperparameter.

We add a task instruction to each plain query (target), resulting in an instruction-condition query $q_{\text{inst}}=\texttt{[IM\_TOKEN] \{task instruction\} \{q\}}$, where \texttt{[IM\_TOKEN]} is replaced by visual tokens. To ensure robustness to linguistic variation, we use a set of ten instruction templates per task~\cite{luo2024skysensegpt}. For example, a text-to-image retrieval instruction might read "Retrieve a satellite image depicting \textit{a coastal city at sunset}", while a visual grounding prompt could be "Identify the object in the given bounding box $[10,25,38,52]$". During training, a template is randomly assigned for each example, teaching the model to follow diverse phrasings of equivalent instructions. 
In Appendix~\ref{sec:inst_ablation}, we evaluate this choice using both known and unseen instructions; diverse-template training is substantially more robust to unseen wording than single-template training.

\subsection{Multimodal Training Data}
To effectively train VLM2GeoVec, we curate a diverse corpus of
approximately \textbf{1.45M multimodal instruction-conditioned contrastive pairs}
from 2.87M available derived examples across public RS datasets, including GeoChat-Instruct \cite{Kuckreja2024GeoChat}, TeoChatlas~\cite{Irvin2024TEOChat}, FIT-RS~\cite{luo2024skysensegpt}, and SkyScript~\cite{wang2024skyscript}. We adapt existing datasets to retrieval by creating new instruction-conditioned query-target contrastive pairs aligned with the RSMEB meta-tasks (see Sec.~\ref{sec:rsmeb}): in scene classification, an image is paired with its class label; in cross‑modal retrieval, images and captions form bidirectional pairs; and image-question pairs with correct answers for VQA. For interleaved tasks, we leverage data with spatial annotations from generative training pipelines~\cite{Kuckreja2024GeoChat,luo2024skysensegpt}, e.g., grounded-image captioning and region-based captioning, to create new pairs for grounded text-to-image retrieval, region-caption retrieval, and region-based composed image retrieval (rCIR). In particular, we extract regions of interest to generate contrastive targets and queries for referring expressions and rCIR, respectively. Finally, we leverage geo-tagged image-text pairs from SkyScript~\cite{wang2024skyscript} for geo-localized tasks. A comprehensive list of training tasks, data sources, and statistics is provided in Appendix~\ref{sec:training_data_extra}.

%% file: experiments.tex
\subsection{Training Details}
We initialize our model from pre‑trained VLM2Vec checkpoints, using Qwen2‑VL as the VLM backbone, based on a ViT‑L-14 image tokenizer. The rationale for this choice of backbone is discussed in Appendix~\ref{sec:training_data_extra}.
We follow a similar training recipe as in~\cite{Jiang2024VLM2Vec}. Fine‑tuning is performed using LoRA with rank $r$$=$$8$ and $\alpha$$=$$64$ (10,422,272 params). We fix the contrastive‐loss temperature at 0.02 and truncate multimodal inputs to 4,096 tokens. The training proceeds for 2,000 steps at a uniform image resolution of $336\times336$. We use GradCache~\cite{gao2021gradcache} to accumulate gradients across sub‑batches (size 6 and 3 for the 2B and 7B model) to reach an effective global batch size of 1,024. We employ AdamW optimizer~\cite{loshchilov2017adamw} with an initial learning rate of $2\times10^{-5}$. We warm up the learning rate linearly over the first 200 steps, then decay it according to a cosine schedule. To balance task contributions, we cap each task subset at 100K examples, reducing the 2,871,323 available pairs to 1,454,119 training pairs, following the multi-task balancing strategy of~\cite{Jiang2024VLM2Vec}. All experiments are conducted on a single node with up to eight NVIDIA A100 80GB GPUs for 1 to 2 days.

\subsection{Baselines}
We compare VLM2GeoVec with two categories of contemporary models: general-purpose VLMs, including CLIP~\cite{radford2021clip} and the universal embedder VLM2Vec~\cite{Jiang2024VLM2Vec} (a contrastive VLM that also follows instructions), and specialized RS dual‑encoders, including RemoteCLIP \cite{liu2024remoteclip}, SkyCLIP \cite{wang2024skyscript}, and GeoRSCLIP~\cite{zhang2024georsclip}, as well as the generative assistant GeoChat \cite{Kuckreja2024GeoChat}. If available, baselines use the ViT-L-14 visual backbone for a fair comparison. On multimodal benchmarks, we employ score-level fusion for dual-encoder baselines, combining the modality-specific feature vectors element-wise with equal weights~\cite{wei2024uniir}. We omit instructions for these baselines, since adding instructions typically hurts the performance of models trained without them~\cite{Jiang2024VLM2Vec}. \textbf{We evaluate 7B VLM variants} for GeoChat, VLM2Vec, and VLM2GeoVec, unless stated otherwise. A comprehensive set of inference task prompts can be found in Appendix~\ref{sec:instructions_extra}.
Evaluation samples are training-disjoint, but task exposure differs across methods; results therefore reflect end-to-end capability rather than architecture alone (see Appendix~\ref{sec:evaluation_data_extra}).

\subsection{Main Results}

\begin{table*}[t]
\caption{Zero-shot classification evaluation on RS datasets. Columns show accuracy (\%) for each dataset, Friedman ranking score, and final rank. GeoRSCLIP is evaluated in-distribution for Million-AID. All embedding models use a 20-prompt ensemble.
}
\centering
\small
\resizebox{0.95\linewidth}{!}{
\begin{tabular}{lcccccccc}
\toprule
Method                                      & AID   & Million-AID& RSI-CB& EuroSAT & UCM & PatternNet & Score & Rank \\
\midrule
CLIP                                                  & 70.10 & 62.24      & 40.25    & 29.30   & 75.76    & 70.76      & 4.8           & 6    \\
VLM2Vec                                            & 64.25 & 58.92      & 32.23    & 21.26   & 69.67    & 62.96      & 6.5           & 7    \\
RemoteCLIP                                            & \underline{75.35} & 49.48      & \underline{51.44}    & 26.67   & \textbf{91.38} & 60.07  & 4.2           & 4    \\
SkyCLIP                                               & 71.75 & \textbf{67.55} & \textbf{52.62} & \textbf{55.63} & 77.71 & \underline{78.15}  & \underline{2.5}           & \underline{2}    \\
GeoRSCLIP                                             & 72.85 & \underline{65.54}      & 51.26    & \underline{51.15}   & 78.10    & 76.35      & 3.0           & 3    \\
\midrule
GeoChat                                            & 73.55 & 57.78      & 44.35    & 36.56   & 84.43    & 64.09      & 4.3           & 5    \\
\midrule
\rowcolor{LightCyan}
\textbf{VLM2GeoVec}                                & \textbf{77.25} & 64.82      & 44.54    & 39.89   & \underline{90.24}    & \textbf{79.76} & \textbf{2.3}           & \textbf{1}    \\
\bottomrule
\end{tabular}
}
\label{tab:classification_evaluation_ensemble}

\end{table*}

\begin{table*}[t]
\centering
\caption{Zero‑shot image-to-text (I2T) and text-to-image (T2I) retrieval evaluation on RSITMD, RSICD, and UCM-caption datasets. Columns show average recall over R@1, R@5, and R@10 for each dataset-task pair, Friedman ranking score, and final rank. RemoteCLIP is evaluated in‑distribution across all tasks.}
\resizebox{0.8\linewidth}{!}{
\begin{tabular}{lcccccccc}
\toprule
  & \multicolumn{2}{c}{RSITMD} 
  & \multicolumn{2}{c}{RSICD} 
  & \multicolumn{2}{c}{UCM-caption} 
  &  &  \\
  \cmidrule(r){2-3}
  \cmidrule(r){4-5}
  \cmidrule(r){6-7}
Method
  & I2T & T2I 
  & I2T & T2I 
  & I2T & T2I 
  & Score & Rank \\
\midrule
CLIP                   & 26.33 & 35.22 & 20.16 & 23.03 & 41.43 & 47.37 & 5.0  & 5 \\
VLM2Vec             & 29.27 & 34.62 & 23.64 & 25.90 & 42.86 & 49.59 & 4.2  & 4 \\
SkyCLIP                & 26.03 & 32.21 & 17.69 & 20.62 & 43.49 & 46.35 & 5.7  & 6 \\
GeoRSCLIP              & \underline{35.69} & 38.30 & \underline{27.57} & 26.75 & 45.71 & 52.26 & 2.7  & 3 \\
\midrule
RemoteCLIP             & \textbf{43.21} & \textbf{48.97} & \textbf{33.61} & \textbf{34.73} & \textbf{52.86} & \textbf{57.81} & \textbf{1.0}  & \textbf{1} \\
\midrule
\rowcolor{LightCyan}
\textbf{VLM2GeoVec} & 31.86 & \underline{41.03} & 21.38 & \underline{28.31} & \underline{48.09} & \underline{52.76} & \underline{2.5}  & \underline{2} \\
\bottomrule
\end{tabular}
}
\label{tab:retrieval_evaluation_compact}
\end{table*}

\begin{table*}[t]
\caption{VQA evaluation on LRBEN and HRBEN datasets. Columns show Precision@1 (\%) for each dataset, Friedman ranking score, and final rank. GeoChat and VLM2GeoVec are evaluated in-distribution across LRBEN tasks. $\ddagger$: results copied from \cite{Kuckreja2024GeoChat}.}
\centering
\small
\resizebox{0.95\linewidth}{!}{
\begin{tabular}{lccccccc}
\toprule 
  & \multicolumn{3}{c}{LRBEN} 
  & \multicolumn{2}{c}{HRBEN} &  &  \\
    \cmidrule(r){2-4}
    \cmidrule(r){5-6}
Method
  & Presence & Comparison & Rural-Urban 
  & Presence & Comparison & Score & Rank \\
\midrule
CLIP                          & 75.03          & 33.26            & 68.00             & 39.18          & 33.38             & 5.1  & 5 \\
VLM2Vec                       & 62.03          & 38.40            & 44.00             & \underline{59.12}          & 43.11             & 4.2  & 3 \\
RemoteCLIP                    & 75.03	       & 33.26            & 44.00             & 39.18          & 33.41             & 5.4  & 7 \\
SkyCLIP                       & 75.03          & 33.26            & 46.00             & 39.18          & 33.38             & 5.3  & 6 \\
GeoRSCLIP                     & 75.03          & 33.26            & 44.00             & 39.50          & 33.69             & 4.8  & 4 \\
\midrule
GeoChat$^\ddagger$               & \textbf{91.09} & \textbf{90.33}   & \textbf{94.00}    & 58.45          & \textbf{83.19}    & \textbf{1.5}  & \textbf{1} \\
\midrule
\rowcolor{LightCyan}
\textbf{VLM2GeoVec}        & \underline{89.78}   & \textbf{90.33}  &  \underline{86.00}  & \textbf{69.47} &   \underline{79.81}  & \underline{1.7}  & \underline{2} \\
\bottomrule
\end{tabular}
}
\label{tab:vqa_evaluation}
\end{table*}

\begin{table*}[t]
\caption{Other multimodal retrieval evaluations.  
Columns show precision@1 (\%) for region-based composed image retrieval (rCIR), referring-expression retrieval (RefExp), geo-localized text-to-image retrieval (GeoT2I), region-caption retrieval (RegCap), and grounded text-to-image retrieval (GrT2I). Meta-tasks include: retrieval, visual grounding (VG), semantic reverse geo-localization (SRG), and spatial localization. Friedman score/rank is computed only for multi-task meta-tasks. VLM2GeoVec is evaluated in-distribution across all tasks.}
\centering
\small
\resizebox{0.80\linewidth}{!}{
\begin{tabular}{lccccccc}
\toprule
            & \multicolumn{1}{c}{Retrieval} 
            & \multicolumn{1}{c}{VG} 
            & \multicolumn{1}{c}{SRG} 
            & \multicolumn{4}{c}{Spatial Localization} \\
\cmidrule(r){2-2} \cmidrule(r){3-3} \cmidrule(r){4-4} \cmidrule(r){5-8}
Method      & rCIR  
            & RefExp   
            & GeoT2I 
            & RegCap 
            & GrT2I      
            & Score 
            & Rank \\
\midrule
CLIP                                   & 2.48        & 13.15       & 2.9        & 1.04   & 0.98       & 2.5  & 2 \\
VLM2Vec                                & 1.98        &  6.65       & 4.4        & 1.25   & 0.86       & 3.0  & 3 \\
RemoteCLIP                             & 1.87        &  4.35       & 1.6        & 1.00   & 0.61       & 4.5  & 5 \\
SkyCLIP                                & 3.96        & 11.85       & 5.1        & 0.64   & 0.92       & 4.0  & 4 \\
GeoRSCLIP                              & 1.16        &  8.25       & 2.0        & 0.46   & 0.49       & 6.0  & 6 \\
\midrule
\rowcolor{LightCyan}
\textbf{VLM2GeoVec}                 & \textbf{22.99} & \textbf{32.50} & \textbf{17.80} & \textbf{26.56} & \textbf{13.70} & \textbf{1.0}  & \textbf{1} \\
\bottomrule
\end{tabular}
}
\label{tab:grounding_evaluation}
\end{table*}

\begin{table*}[t]
\caption{Evaluation summary on RSMEB. Rows show the Friedman score and final rank across all tasks.}
\centering
\small
\resizebox{0.95\linewidth}{!}{
\begin{tabular}{lccccccc}
\toprule
       & CLIP & VLM2Vec & RemoteCLIP & SkyCLIP & GeoRSCLIP & GeoChat & VLM2GeoVec \\
\midrule
\rowcolor{LightCyan}
Score                 & 4.57  & 4.68  & 3.81  & 4.11  & 3.86  & 3.04  & \textbf{1.93} \\
\rowcolor{LightCyan}
Rank                  & 6     & 7     & 3     & 5     & 4     &  2     & \textbf{1}     \\
\bottomrule
\end{tabular}
}
\label{tab:full_evaluation_results}
\end{table*}

We introduce six meta-tasks: classification, retrieval, visual question answering, visual grounding, spatial localization, and semantic reverse geo‑localization. We evaluate each task using metrics according to the literature: accuracy for classification; recall@1/5/10 (R@k) and the average of the three metrics for cross‑modal retrieval~\cite{liu2024remoteclip,wang2024skyscript}; and precision@1 (P@1) for other tasks, similar to MMEB~\cite{Jiang2024VLM2Vec}. 

We use a ranking-based evaluation for a consistent comparison across diverse settings~\cite{wang2022usb,SanchezAimar2024ADELLO_LTSSL}: for each dataset we rank methods (best$=$1), then compute the average rank over all datasets (Friedman score)~\cite{Friedman1937ranking,Friedman1940ranking}, \text{score}$=$$\frac{1}{m}\sum_{i=1}^{m}\text{rank}_i$ (lower is better) and finally derive the order from the scores.

\textbf{Classification.}
Table~\ref{tab:classification_evaluation_ensemble} reports zero-shot accuracy across various classification datasets, where we employ an ensemble of 20 prompts for a fair comparison with embedding methods~\cite{radford2021clip,wang2024skyscript,zhang2024georsclip}. We observe that VLM2GeoVec significantly outperforms general-purpose baselines, e.g., CLIP, and VLM2Vec, delivering improvements over VLM2Vec ranging from approximately +5.9 to +16.8 percentage points (pp) on Million-AID and PatternNet, respectively, confirming that instruction-conditioned domain adaptation substantially enhances visual discrimination. Generally, VLM2GeoVec exhibits competitive performance, ranking first in zero-shot classification and surpassing all specialized RS dual encoders in AID and PatternNet. This indicates that our approach generalizes well to unseen datasets for scene classification and can compensate for larger RS cross-modal corpora, e.g., 5M~\cite{wang2024skyscript,zhang2024georsclip} vs.~our 0.55M, through instruction-aware multi-task training.

\textbf{Retrieval.}
In zero‑shot image-text (I2T) and text-image (T2I) retrieval (Table~\ref{tab:retrieval_evaluation_compact}), we report the mean of R@1/5/10 for each dataset-task pair and rank methods by Friedman score. VLM2GeoVec ranks second, trailing only RemoteCLIP, which was trained on RSITMD+RSICD+UCM-caption data, while leading all other zero-shot baselines. We observe the most pronounced gains on text-to-image tasks, with improvement increases ranging from 0.5 pp to 2.7 pp compared to the best remote-sensing foundation model, GeoRSCLIP, demonstrating strong generalization for cross-modal retrieval.

CIR measures the model’s ability to retrieve a target image when given a source image and a text modification that describes how to alter it~\cite{vo2019composing,Liu_2021CIRR}. In the region-based variant (rCIR), the source is a cropped region from a larger scene paired with text that references and extends beyond that region. Retrieval then requires grounding the region and extrapolating context changes to recover the full image. We evaluate rCIR using precision@1 (Table~\ref{tab:grounding_evaluation}). VLM2GeoVec achieves 22.99\%, more than five times higher than the best specialized dual encoder baseline, SkyCLIP (3.96\%), and almost ten times higher than general-purpose CLIP (2.48\%). Our unified multimodal encoder with deep modality integration significantly outperforms simpler score-fusion approaches, demonstrating effective grounding of region-level inputs and composition with text-driven contextualization. In comparison, VLM2Vec, which has been trained on everyday CIR data but lacks RS adaptation, struggles to reconcile the unique spatial and spectral characteristics of satellite imagery, such as large variations in scale, resulting in substantially lower precision in image-text compositionality.

\textbf{Visual Question Answering.}
We evaluate VLM2GeoVec on the LRBEN and HRBEN VQA benchmarks, reporting precision@1 (P@1) for presence, comparison, and rural-urban question types (Table~\ref{tab:vqa_evaluation}). We observe that VLM2GeoVec outperforms all embedding baselines by large margins of 19-50 pp on HRBEN (zero‑shot), highlighting the intrinsic limitations of late-fusion methods, which shows the benefits of deep image-question multimodal integration for RS data. Conversely, VLM2Vec, which has been trained on multiple VQA datasets, still fails to provide the correct answer to RS questions in most scenarios. Although GeoChat, a generative assistant, naturally excels in VQA—recording above 90\% P@1 on LRBEN presence and rural-urban questions—VLM2GeoVec still manages to deliver solid performance in VQA, trailing GeoChat by only 1.3\,pp on LRBEN presence, matching on LRBEN comparison, and even surpassing the generative assistant on high‑resolution HRBEN presence detection questions by 11 pp. These results suggest that instruction-conditioned contrastive pretraining could close the gap with generative assistants and excel at a variety of multiple-choice questions, which require spatial detail extraction in RS contexts.

\textbf{Visual Grounding.} 
This task is assessed via referring-expression retrieval (RefExp), where the model must identify the correct region given a textual query (Table~\ref{tab:grounding_evaluation}). VLM2Vec, despite instruction conditioning on the same task, was trained on natural images and achieves only 6.65\% P@1, struggling to match referring expressions in the remote-sensing domain. Simple dual-encoder baselines that interpolate image and text features on robust RS models marginally surpass this performance. In contrast, VLM2GeoVec’s interleaved modality design with domain-adapted LoRA adapters yields 32.5\% P@1, and even our 2B variant reaches 21.55\%, outstripping all baselines and demonstrating the effectiveness of domain adaptation combined with instruction-conditioned contrastive learning.

\textbf{Spatial Localization.}
This capability is evaluated with two tasks: region-caption retrieval (RegCap), which retrieves a region caption given a full-scene image and a query bounding box, and grounded text-to-image retrieval (GrT2I), which retrieves the correct full-scene image given a text prompt that includes bounding-box annotations. In RegCap, VLM2GeoVec achieves 26.56\% precision@1—up from 1.25\% (+25.31\,pp)—and in GrT2I, 13.7\% precision@1—up from 0.86\% (+12.84\,pp). The instruction-conditioned VLM2Vec backbone, unadapted to spatial prompts, cannot follow these localization instructions. Dual-encoder models plateau in the single digits because they perform late fusion of separate image and text embeddings and fail to integrate spatial bounding boxes with language. By contrast, our interleaved image, text, and spatial coordinates combined with instruction-conditioned contrastive pretraining deliver end-to-end grounding and retrieval performance, capabilities vital for real-world applications such as disaster assessment and land-cover mapping.

\textbf{Semantic Reverse Geo-localization.}
Geo-localized text-to-image retrieval (GeoT2I) evaluates the task where the input is a caption with embedded latitude-longitude coordinates and the output is the matching satellite image. We report precision@1 for GeoT2I. VLM2GeoVec achieves 17.8\% precision@1, compared to 2\% for GeoRSCLIP and 5.1\% for SkyCLIP (Table~\ref{tab:grounding_evaluation}). These results show that embedding spatial metadata directly into the token stream yields robust semantic alignment in RS applications. Note that SkyCLIP, albeit trained on SkyScript images, cannot leverage geo-coordinates to refine retrieval, underscoring the unique advantage of our interleaved geo-token approach.
\subsection{Discussion}
\label{sec:discussion_ablations}

\begin{table*}[t]
\caption{Ablation of initialization and model capacity. Strategies in columns. The Friedman score is the average rank across all RSMEB tasks; the final rank is determined by this score; lower is better. Models \textit{trained from scratch} use LoRA over base VLM ((\textbf{Q})wen2-VL), instead of boostrapping from contrastively-pretrained weights (VLM2(\textbf{V})ec). Model sizes: 2B/7B.}
\centering
\small
\resizebox{0.70\linewidth}{!}{
\begin{tabular}{lcccccc}
\toprule
  & \multicolumn{2}{c}{VLM2Vec} 
  & \multicolumn{4}{c}{VLM2GeoVec} 
  \\
  \cmidrule(r){2-3}
  \cmidrule(r){4-7}
                      &  2B (\textbf{Q})  &    7B (\textbf{Q}) &    2B (\textbf{Q}) &     2B   (\textbf{V})         & 7B (\textbf{Q})  &      7B  (\textbf{V})     \\
\midrule
\rowcolor{LightCyan} 
Score & 4.86 & 4.59 & 4.59 & 3.14 & 2.45 & \textbf{1.36} \\
\rowcolor{LightCyan}
Rank  & 6      & 4      & 4      & 3      & 2      & \textbf{1}      \\
\bottomrule
\end{tabular}
}
\label{tab:ablation_results}
\end{table*}

\comment{

\begin{table*}[t]
\caption{Ablation of instruction-tuning and phrasing robustness. Strategies in columns. The Friedman score is the average rank across all RSMEB tasks; the final rank is determined by this score; lower is better.}
\centering
\small
\resizebox{0.8\linewidth}{!}{
\begin{tabular}{lcccc}
\toprule
 & diverse-known & single-known & diverse-unseen & single-unseen \\
\midrule
\rowcolor{LightCyan} 
Score & \textbf{2.16} & 2.27 & 2.34 & 3.23 \\
\rowcolor{LightCyan} 
Rank  & \textbf{1}    & 2    & 3    & 4    \\
\bottomrule
\end{tabular}
}
\label{tab:inst_ablation}
\end{table*}

}

\textbf{Overall results.}
Table \ref{tab:full_evaluation_results} groups seven methods into three categories: general-purpose VLMs (CLIP and VLM2Vec), specialized remote-sensing dual‑encoders (RemoteCLIP, SkyCLIP, and GeoRSCLIP), and the instruction‑tuned assistant GeoChat. VLM2GeoVec achieves the lowest overall Friedman score (1.93), outperforming all baselines by a substantial margin and demonstrating that instruction‑conditioned contrastive pretraining more effectively aligns multimodal tasks by integrating domain-specific remote‑sensing features with textual prompts. GeoChat’s generative framework excels on VQA benchmarks, yet it lacks the scalable retrieval capabilities needed for full RSMEB coverage. The specialized dual‑encoders improve over general‑purpose VLMs but still cannot match VLM2GeoVec’s multi‑task capabilities. Interestingly, VLM2Vec (4.86) underperforms CLIP (4.59), which may reflect CLIP’s broader, large‑scale pretraining. Overall, domain‑adapted pretraining yields substantial improvements compared to the general‑purpose instruction-following VLM baseline and instruction-free specialized dual-encoders, underscoring its significant impact on both retrieval and reasoning performance.

\textbf{Ablation studies.}
We first assess two key design choices—\textbf{initialization strategy} and \textbf{model capacity}—using Friedman scores and ranks across RSMEB (Table \ref{tab:ablation_results}). Bootstrapping from contrastively pre-trained VLM2Vec weights cuts the Friedman score of the 7B model from 2.45 to 1.36, showing that instruction-conditioned pretraining provides transferable multimodal priors that accelerate convergence and strengthen multimodal alignment. Nevertheless, \textit{VLM2Vec pretrained weights are not a hard dependency}, and VLM2GeoVec can also be initialized directly from Qwen2-VL and then trained exclusively on remote-sensing data to achieve competitive results. Increasing capacity from 2B to 7B consistently boosts performance: our VLM2GeoVec-7B, initialized from VLM2Vec weights, outperforms its 2B counterpart, and even our 7B model trained from scratch surpasses the bootstrapped 2B variant. Finally, under comparable compute and time budgets, our remote-sensing-tailored, instruction-conditioned contrastive pretraining from scratch delivers embeddings on par with—or better than—those from the general-purpose VLM2Vec, with the scratch-initialized VLM2GeoVec-2B matching the general-purpose VLM2Vec-7B’s performance. These results underscore that combining larger VLMs with domain-adapted pretraining yields richer, more general embeddings for diverse retrieval and reasoning challenges in RS. A comprehensive set of quantitative results, ablations, and qualitative examples can be found in Appendices~\ref{sec:main_results_extra}--\ref{sec:qualitative_results_extra}.


%% file: conclusion.tex
We propose RSMEB, a novel multimodal retrieval benchmark, spanning six meta-tasks---classification, multimodal retrieval, VQA, visual grounding, spatial localization, and semantic geo-localized retrieval---with a focus on assessing interleaved modality integration. We also introduce VLM2GeoVec, a unified, instruction-following multimodal embedding model that interleaves images, text, bounding boxes, and geographic coordinates into a shared embedding space, trained contrastively with lightweight LoRA adapters. We show that our improved baseline unlocks interleaved retrieval for remote sensing, achieving up to +25 pp in region-caption retrieval, +19 pp in referring-expression retrieval, and over $3\times$ improvement in geo-T2I, while matching or exceeding competitive baselines on conventional tasks.

\textbf{Limitations.}
VLM2GeoVec resamples imagery to $336\times336$ for compatibility with the visual tokenizer and feasible contrastive training, which may remove small-object detail in very-high-resolution scenes. The current model is restricted to single-view RGB imagery, text, and textual metadata, without native support for multi-temporal data or additional RS modalities (e.g., SAR, multispectral imagery, or LiDAR)~\cite{xiong2025dofaclip}. Moreover, encoding coordinates as text provides only coarse geographic context and does not explicitly model continuous spatial relationships or local geographic structure~\cite{klemmer2025satclip}.

\textbf{Future Work.}
Future work could incorporate richer spatial representations, such as learned geodesic embeddings, and extend the unified encoder to additional RS modalities~\cite{xiong2025dofaclip}, temporal modeling~\cite{Irvin2024TEOChat}, and segmentation~\cite{Wei2020HRSID_segmentation,garioud2023flair}. Tiled encoding, region crops, multi-scale processing, and native-resolution tokenization are promising directions for handling very-high-resolution imagery.

%% file: acknowledgements.tex
This work was supported by the Wallenberg Artificial Intelligence, Autonomous Systems and Software Program (WASP), funded by the Knut and Alice Wallenberg Foundation, and the Swedish Research Council with grant agreement no. 2016-05543, no. 2022-04266 and no. 2024-05652. The computational resources were provided by the National Academic Infrastructure for Supercomputing in Sweden (NAISS), partially funded by the Swedish Research Council through grant agreement no. 2022-06725, and by the Berzelius resource, provided by the Knut and Alice Wallenberg Foundation at the National Supercomputer Centre.

%% file: appendix.tex
The appendix includes the following sections:
\begin{enumerate}
    \item \textbf{Training Dataset} (Appendix~\ref{sec:training_data_extra}): presents additional details about the construction of our training corpus.

    \item \textbf{Benchmark Dataset} (Appendix~\ref{sec:evaluation_data_extra}): presents additional details about the construction of RSMEB.

    \item \textbf{Detailed RSMEB Results} (Appendix~\ref{sec:main_results_extra}): presents the full list of results for the methods under comparison.

    \item \textbf{Detailed Ablation Study} (Appendix~\ref{sec:ablation_results_extra}): presents the full list of results for ablated models.

    \item \textbf{Qualitative Results} (Appendix~\ref{sec:qualitative_results_extra}): presents qualitative results across several tasks.

    \item \textbf{Task Prompts} (Appendix~\ref{sec:instructions_extra}): presents a list of query and target prompts used in this work.
\end{enumerate}

\section{Training Dataset}
\label{sec:training_data_extra}

\textbf{Corpus curation.} Our final pretraining corpus contains \textbf{1,454,119 samples} drawn from several public remote-sensing datasets and organized into 21 task subsets. Before per-task balancing, the source datasets provide \textbf{2,871,323 available pairs}. As detailed in
Table~\ref{tab:dataset_expanded_samples}, each task subset is capped at 100K examples when necessary to prevent large datasets from dominating multi-task training. We cover scene classification (e.g., GeoChat, FIT-RS, TeoChatlas), cross-modal retrieval in both directions, composed image retrieval, visual question answering, referring-expression retrieval, region-caption retrieval, grounded retrieval (both image-to-text and text-to-image), geo-localized semantic retrieval (SkyScript), and image-to-image matching. Each row specifies the source dataset, input/output modalities, and exact sample counts used (with parenthetical totals indicating the full dataset sizes). This diverse mix ensures the model learns from high-resolution and low-resolution imagery, multi-region grounding, and geospatial context, providing a rich foundation for downstream evaluation. Table~\ref{tab:dataset_licenses} details the license of the datasets used in this work for training and evaluation purposes.
 
\begin{table}[t]
  \centering
\caption{Pretraining task subsets, source datasets, modalities, and sample counts. ``Used'' denotes the final balanced training corpus; parenthetical values denote the total number of available source pairs before per-task capping.}
  \label{tab:dataset_expanded_samples}  
  \resizebox{\linewidth}{!}{
  \begin{tabular}{llll}
    \toprule
    \textbf{Multimodal Task}          & \textbf{Source Dataset}   & \textbf{Input $\to$ Output}       & \textbf{\#Used (\#Available)} \\
    \midrule
    Classification                    & GeoChat                  & Image $\to$ Class label            & 31,500 \\
    Classification                    & FIT-RS                   & Image $\to$ Class label            & 100,000 (108,641) \\
    Classification                    & TeoChatlas               & Image $\to$ Class label            & 45,101 \\
    Image-to-text Retrieval           & SkyScript                & Image $\to$ Text                   & 100,000 (379,722) \\
    Text-to-image Retrieval           & GeoChat                  & Text $\to$ Image                   & 88,773 \\
    Text-to-image Retrieval           & FIT-RS                   & Text $\to$ Image                   & 86,956 \\
    Text-to-image Retrieval           & SkyScript                & Text $\to$ Image                   & 100,000 (379,722) \\
    Composed Image Retrieval          & FIT-RS                   & Image + Text $\to$ Image           & 72,026 \\
    Composed Image Retrieval          & TeoChatlas               & Image + Text $\to$ Image           & 68,943 \\
    Visual Question Answering         & GeoChat                  & Image $\to$ Text answer            & 78,053 \\
    Visual Question Answering         & FIT-RS                   & Image $\to$ Text answer            & 100,000 (389,675) \\
    Referring-expression Retrieval    & GeoChat                  & Image + Text $\to$ Image Region    & 64,680 \\
    Region-caption Retrieval          & GeoChat                  & Image + BBox $\to$ Text            & 69,270 \\
    Region-caption Retrieval          & FIT-RS                   & Image + BBox $\to$ Text            & 75,362 \\
    Grounded T2I Retrieval            & GeoChat                  & Text + BBoxes $\to$ Image          & 17,758 \\
    Grounded T2I Retrieval            & FIT-RS                   & Text + BBoxes $\to$ Image          & 49,814 \\
    Grounded I2T Retrieval            & GeoChat                  & Image $\to$ Text + BBoxes          & 17,758 \\
    Grounded I2T Retrieval            & FIT-RS                   & Image $\to$ Text + BBoxes          & 49,814 \\
    Geo-localized T2I Retrieval       & SkyScript                & Text + (Lat,Lon) $\to$ Image       & 100,000 (379,722) \\
    Geo-localized I2T Retrieval       & SkyScript                & Image + (Lat,Lon) $\to$ Text       & 100,000 (379,722) \\
    Image-to-image Retrieval          & TeoChatlas               & Image $\to$ Image                  & 38,311 \\
    \midrule
    \multicolumn{3}{l}{\textbf{Total}}                                           & \textbf{1,454,119} (2,871,323) \\
    \bottomrule
  \end{tabular}
  }
\end{table}

\begin{table}[t]
\centering
\caption{Dataset licenses.}
\label{tab:dataset_licenses}
\begin{tabular}{@{}lp{0.6\linewidth}@{}}
\toprule
\textbf{Dataset} & \textbf{License} \\
\midrule
AID & Uses Google Earth/Maps imagery—governed by Google Earth/Maps ToS. \\
Million‐AID & Not specified. \\
RSI‐CB & CC BY‐NC 4.0. \\
EuroSAT & MIT License. \\
UCM (UCMerced) & No separate license; images from USGS/Google Earth (research‐only use). \\
PatternNet & Not specified; uses Google Earth imagery (see AID). \\
RSITMD & Not specified; academic‐use terms. \\
RSICD & Not specified. \\
UCM‐caption & Follows UCM license (USGS/Google Earth terms). \\
RSVQA  & CC BY 4.0 International. \\
FIT‐RS  & CC BY‐NC 4.0. \\
GeoChat‐Instruct & Apache‐2.0 (according to HuggingFace release). \\
TeoChatlas & Apache‐2.0 (according to HuggingFace release). \\
SkyScript & MIT License. \\
\bottomrule
\end{tabular}
\end{table}

\section{RSMEB dataset}
\label{sec:evaluation_data_extra}

\textbf{Benchmark construction.} While we do not collect new data, we introduce a \emph{simplified ranking protocol} with inputs-targets designed to assess valuable RS retrieval capabilities across diverse tasks and modalities.
\emph{Harmonization:} each task is derived from a single source (no cross-dataset mixing) and is curated by removing missing/invalid labels or region descriptions, tiny bboxes, and near-duplicate image/text targets; by standardizing bboxes and merging/extracting relevant regions (RefExp/rCIR/RegCap). \textit{Candidate pools} are capped at 2K (RefExp/GeoT2I, inspired by MMEB~\cite{Jiang2024VLM2Vec}). \emph{No custom negatives:} we rank against the full task pool.

\textbf{Train-test provenance.} We curated \textbf{RSMEB only from official test splits} (GeoChat-Bench/TeoChatlas-test/FIT-RS-test/SkyScript-retrieval) plus standard classification and cross-modal retrieval datasets (see Table~2 for data sources). GeoChat and TeoChatlas ensure evaluation in val/test splits from the original datasets. \textbf{Our training uses only official train splits} (GeoChat-Instruct: NWPU/LRBEN(train)/FloodNet/DOTA/DIOR/FAIR1M; TeoChatlas: fMoW/xBD/S2Looking/QFabric). For FIT-RS, we \textbf{keep only custom data and remove public datasets overlapping with evaluation} (e.g., UCM / RSITMD classification and datasets already in GeoChat). In general, the train and test data are disjoint by source-defined splits and explicit overlap filtering.

\textbf{Evaluation exposure.} We distinguish sample overlap from task-format and source-family exposure. No evaluation sample is used for training. However, for some interleaved evaluations, VLM2GeoVec is trained on related task formulations or held-out splits from source families represented during training. Thus, sample-disjoint evaluation does not always imply an entirely unseen task format.

\section{Detailed RSMEB results}
\label{sec:main_results_extra}

\subsection{Main Results}

 Figure 1 (main paper) presents a radar-plot overview of each method’s performance across the six benchmark categories (classification, multimodal retrieval, visual grounding, spatial localization, semantic reverse geo-localization, and VQA). For a complete view, Table~\ref{tab:full_evaluation_results_extra} lists per-dataset and per-task metrics—including top-1 accuracy for classification sets, average recall@\{1,5,10\} for cross-modal retrieval, and precision@1 for all other tasks—as well as each method’s overall Friedman score and rank. To examine the retrieval subtasks in finer detail, Table \ref{tab:retrieval_evaluation} breaks down image-to-text and text-to-image performance on RSITMD, RSICD, and UCM-caption, reporting Recall@1, Recall@5, and Recall@10 for every embedding model.

\begin{table}[t]
\caption{Comprehensive evaluation on RSMEB. Methods in columns, datasets/tasks in rows. Rows show top-1 accuracy (\%) for classification datasets, average recall over R@\{1,5,10\} (\%) for cross-modal retrieval tasks, and precision@1 (\%) for other tasks. The Friedman score is the average rank across all tasks (lower is better), and the rank is the position based on that score. GeoChat, VLM2Vec, and VLM2GeoVec are 7B-parameter variants. Methods evaluated in-distribution are highlighted in a yellow background.}
\label{tab:full_evaluation_results_extra}
\centering
\small
\resizebox{\linewidth}{!}{
\begin{tabular}{lccccccc}
\toprule
Dataset        & CLIP & VLM2Vec & RemoteCLIP & SkyCLIP & GeoRSCLIP & GeoChat & VLM2GeoVec \\
\midrule
\rowcolor{Lavender}
\multicolumn{8}{l}{\textbf{Classification}} \\
AID                   & 70.10 & 64.25 & \underline{75.35} & 71.75 & 72.85 & 73.55 & \textbf{77.25} \\
Million-AID           & 62.24 & 58.92 & 49.48 & \textbf{67.55} & \cellcolor{LightYellow}\underline{65.54} & 57.78 & 64.82 \\
RSI-CB              & 40.25 & 32.23 & \underline{51.44} & \textbf{52.62} & 51.26 & 44.35 & 44.54 \\
EuroSAT               & 29.30 & 21.26 & 26.67 & \textbf{55.63} & \underline{51.15} & 36.56 & 39.89 \\
UCM                   & 75.76 & 69.67 & \textbf{91.38} & 77.71 & 78.10 & 84.43 & \underline{90.24} \\
PatternNet            & 70.76 & 62.96 & 60.07 & \underline{78.15} & 76.35 & 64.09 & \textbf{79.76} \\
\midrule
\rowcolor{Lavender}
\multicolumn{8}{l}{\textbf{Retrieval}} \\
RSITMD I2T            & 26.33 & 29.27 & \cellcolor{LightYellow}\textbf{43.21} & 26.03 & \underline{35.69} & –     & 31.86 \\
RSITMD T2I            & 35.22 & 34.62 & \cellcolor{LightYellow}\textbf{48.97} & 32.21 & 38.30 & –     & \underline{41.03} \\
RSICD I2T             & 20.16 & 23.64 & \cellcolor{LightYellow}\textbf{33.61} & 17.69 & \underline{27.57} & –     & 21.38 \\
RSICD T2I             & 23.03 & 25.90 & \cellcolor{LightYellow}\textbf{34.73} & 20.62 & 26.75 & –     & \underline{28.31} \\
UCM-caption I2T       & 41.43 & 42.86 & \cellcolor{LightYellow}\textbf{52.86} & 43.49 & 45.71 & –     & \underline{48.09} \\
UCM-caption T2I       & 47.37 & 49.59 & \cellcolor{LightYellow}\textbf{57.81} & 46.35 & 52.26 & –     & \underline{52.76} \\
rCIR                  & 2.48  & 1.98  & 1.87  & \underline{3.96}  & 1.16  & –     & \cellcolor{LightYellow}\textbf{22.99} \\
\midrule
\rowcolor{Lavender}
\multicolumn{8}{l}{\textbf{Spatial Localization}} \\
RegCap                & 1.04  & 1.25  & 1.00  & 0.64  & 0.46  & –     & \cellcolor{LightYellow}\textbf{26.56} \\
GrT2I                 & 0.98  & 0.86  & 0.61  & 0.92  & 0.49  & –     & \cellcolor{LightYellow}\textbf{13.70} \\
\midrule
\rowcolor{Lavender}\multicolumn{8}{l}{\textbf{Visual Grounding}} \\
RefExp                & 13.15 & 6.65 &  4.35  & 11.85 & 8.25  & –     & \cellcolor{LightYellow}\textbf{32.50} \\
\midrule
\rowcolor{Lavender}
\multicolumn{8}{l}{\textbf{Semantic Reverse Geo-localization}} \\
GeoT2I                & 2.90  & 4.40  & 1.60  & \underline{5.10}  & 2.00  & –     & \cellcolor{LightYellow}\textbf{17.80} \\
\midrule
\rowcolor{Lavender}
\multicolumn{8}{l}{\textbf{VQA}} \\
LRBEN Presence        & 75.03 & 62.03 & 75.03 & 75.03 & 75.03 & \cellcolor{LightYellow} \textbf{91.09} & \cellcolor{LightYellow} \underline{89.78} \\
LRBEN Comparison      & 33.26 & 38.40 & 33.26 & 33.26 & 33.26 & \cellcolor{LightYellow}\textbf{90.33} & \cellcolor{LightYellow}\textbf{90.33} \\
LRBEN Rural–Urban     & 68.00 & 44.00 & 44.00 & 46.00 & 44.00 & \cellcolor{LightYellow}\textbf{94.00} & \cellcolor{LightYellow}\underline{86.00} \\
HRBEN Presence        & 39.18 & \underline{59.12} & 39.18 & 39.18 & 39.50 & 58.45 & \textbf{69.47} \\
HRBEN Comparison      & 33.38 & 43.11 & 33.41 & 33.38 & 33.69 & \textbf{83.19} & \underline{79.81} \\
\midrule
\rowcolor{LightCyan}
Score                 & 4.57  & 4.68  & 3.81  & 4.11  & 3.86  & 3.04  & \textbf{1.93} \\
\rowcolor{LightCyan}
Rank                  & 6     & 7     & 3     & 5     & 4     &  2     & \textbf{1}     \\
\bottomrule
\end{tabular}
}
\end{table}

\begin{table}[t]
\caption{Zero-shot image–to–text (I2T) and text–to–image (T2I) retrieval evaluations on RSITMD, RSICD, and UCM-caption (UCM) datasets. Reported Recall@1, Recall@5, Recall@10 across the six retrieval scores. RemoteCLIP is evaluated in‑distribution for RSITMD, RSICD, and UCM datasets (highlighted with a gray background).
}
\label{tab:retrieval_evaluation}
\centering
\small
\resizebox{\linewidth}{!}{
\begin{tabular}{llcccccc}
\toprule
Dataset   & Method                             & I2T@1 & I2T@5 & I2T@10 & T2I@1 & T2I@5 & T2I@10   \\
\midrule
RSITMD    & CLIP                               & 11.95 & 28.32 & 38.72  & 14.03 & 38.41 & 53.23  \\
RSITMD    & VLM2Vec-2B                         & 13.05 & 28.76 & 39.38  & 12.70 & 33.94 & 47.08  \\
RSITMD    & VLM2Vec-7B                         & 13.05 & 31.19 & 43.58  & 14.20 & 36.77 & 52.88  \\
RSITMD    & SkyCLIP                            & 12.39 & 27.21 & 38.50  & 12.12 & 33.10 & 51.42  \\
RSITMD    & GeoRSCLIP                          & 19.69 & 38.50 & 48.89  & 17.21 & 41.37 & 56.33  \\
\rowcolor{Gray}
RSITMD    & RemoteCLIP   & 23.23 & 45.35 & 61.06  & 21.42 & 54.12 & 71.37  \\

\midrule
\rowcolor{LightCyan}
RSITMD    & \textbf{VLM2GeoVec-2B}             & 10.62 &  30.97 & 43.81 &  12.43 &  38.81 &  54.38 \\
RSITMD    & \textbf{VLM2GeoVec-2B} from scratch & 8.85    & 26.99   & 40.93    & 11.33   & 36.95   & 52.43  \\

\rowcolor{LightCyan}
RSITMD    & \textbf{VLM2GeoVec-7B}             & 14.82 & 33.85 & 46.90 & 17.48 & 45.04 & 60.58  \\
RSITMD    & \textbf{VLM2GeoVec-7B} from scratch & 13.27 & 30.75   & 43.14    & 15.31   & 40.40   & 56.77   \\

\midrule
RSICD      & CLIP                              &  8.51 & 20.04 & 31.93  &  7.78 & 24.57 & 36.74  \\
RSICD     & VLM2Vec-2B                         &  7.69 & 19.95 & 28.64  &  5.78 & 18.26 & 29.22  \\
RSICD     & VLM2Vec-7B                         & 10.80 & 24.89 & 35.22  &  9.22 & 28.31 & 40.16  \\
RSICD     & SkyCLIP                            &  7.23 & 18.12 & 27.72  &  6.09 & 22.03 & 33.74  \\
RSICD     & GeoRSCLIP                          & 12.72 & 28.82 & 41.17  &  9.97 & 28.18 & 42.10  \\
\rowcolor{Gray}
RSICD     & RemoteCLIP   & 15.55 & 35.77 & 49.50  & 12.90 & 38.02 & 53.27  \\
\midrule
\rowcolor{LightCyan}
RSICD     & \textbf{VLM2GeoVec-2B}             & 
7.78 & 21.32 & 30.56 &  7.30 & 25.45 & 39.45 \\
RSICD     & \textbf{VLM2GeoVec-2B} from scratch & 8.69    & 20.86   & 30.10    & 7.52    & 23.51   & 37.02     \\

\rowcolor{LightCyan}
RSICD     & \textbf{VLM2GeoVec-7B}             &  9.52 & 23.06 & 31.56  &  10.10 & 30.03 & 44.79  \\
RSICD     & \textbf{VLM2GeoVec-7B} from scratch &  8.14    & 22.60   & 33.21    & 8.45  &  27.48   & 41.57  \\

\midrule
UCM       & CLIP                               & 12.38 & 44.29 & 67.62  & 12.29 & 50.19 & 79.62  \\
UCM       & VLM2Vec-2B                         & 13.81 & 48.10 & 70.00  & 13.14 & 46.57 & 76.76  \\
UCM       & VLM2Vec-7B                         &  6.19 & 43.81 & 68.57  & 14.86 & 51.90 & 82.00   \\
UCM       & SkyCLIP                            & 12.86 & 45.71 & 71.90  & 11.14 & 47.90 & 80.00   \\
UCM       & GeoRSCLIP                          & 18.57 & 46.19 & 72.38  & 16.10 & 54.48 & 86.19   \\
\rowcolor{Gray}
UCM       & RemoteCLIP  & 17.62 & 59.05 & 81.90  & 17.43 & 62.38 & 93.62  \\
\midrule
\rowcolor{LightCyan}
UCM       & \textbf{VLM2GeoVec-2B}             &   15.71 & 46.19 & 73.81 & 14.48 & 50.38 & 82.00  \\
UCM       & \textbf{VLM2GeoVec-2B} from scratch & 13.33   & 43.81   & 68.57    & 13.33   & 50.19   & 84.19    \\

\rowcolor{LightCyan}
UCM       & \textbf{VLM2GeoVec-7B}             &   17.14 & 51.90 & 75.24  & 15.62 & 54.67 & 88.00 \\
UCM       & \textbf{VLM2GeoVec-7B} from scratch &  14.29   & 45.71   & 74.29    & 14.00   & 54.57   & 86.57  \\

\bottomrule
\end{tabular}
}
\end{table}

\subsection{Extended Comparison}
\label{sec:extended_comparison}
\begin{figure*}[t]
  \centering
    \centering
    \includegraphics[trim={0.3cm 0.3cm 0.3cm 0.3cm},clip,width=0.7\linewidth]{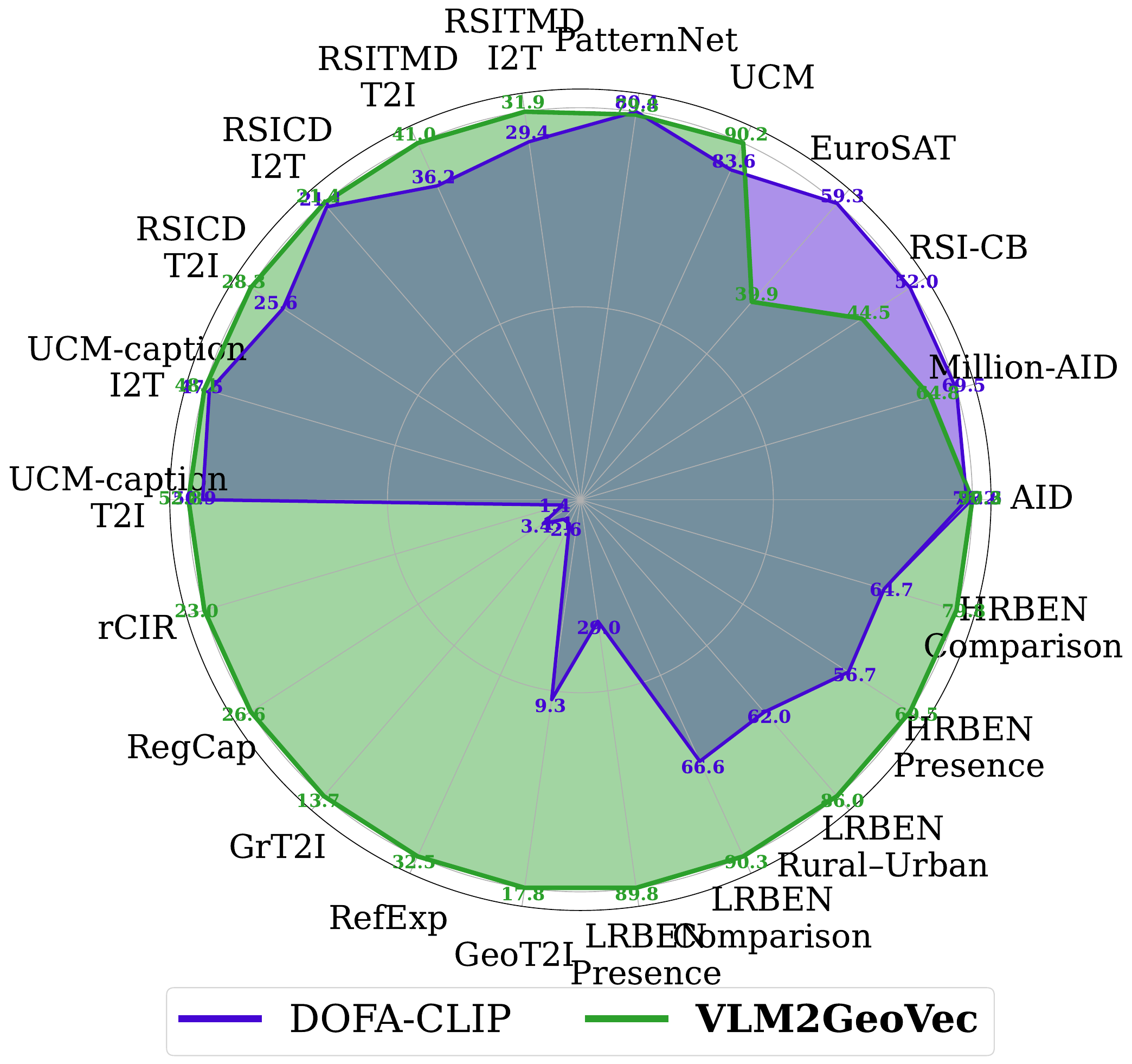}
    \caption{Extended comparison on RSMEB: DOFA-CLIP vs  VLM2GeoVec (ours).}
    \label{fig:comparison_dofa_clip}    
 \end{figure*}

In this section, we compare our work with DOFA-CLIP~\cite{xiong2025dofaclip}, a concurrent remote-sensing dual encoder that supports various multispectral modalities (e.g., SAR and Infrared) in addition to RGB and text. Figure~\ref{fig:comparison_dofa_clip} shows that \textcolor{violet}{\textbf{DOFA-CLIP}} is mainly competitive in cross-modal tasks (e.g., classification), while \textcolor{ForestGreen}{\textbf{VLM2GeoVec}} is consistently stronger in interleaved tasks (e.g., VQA). We attribute this trend to the lack of cross-modal attention (using late fusion) and the absence of contrastive interleaved training in \textcolor{violet}{\textbf{DOFA-CLIP}}, which focuses exclusively on cross-modal alignment. 

\paragraph{Scope of comparisons.}
We compare against a representative set of general-purpose embedders, RS-specific dual encoders, and GeoChat for classification and VQA. Nevertheless, our evaluation is not intended to be exhaustive with respect to
recent generative models, general-purpose multimodal embedding methods, or single-task RS approaches, including composed image retrieval~\cite{psomas2024composed} and visual grounding
~\cite{sun2022visualgrounding}. Accordingly, the reported results should be interpreted relative to the evaluated baselines and as evidence of the breadth of a unified embedding interface across heterogeneous query--target
configurations, rather than as establishing state-of-the-art performance on
each individual task.

\section{Detailed Ablation Results}
\label{sec:ablation_results_extra}

\subsection{Initialization and Model Capacity}

\textbf{Backbone selection.} We use Qwen2-VL because it provides an open multimodal backbone that natively supports interleaved image and text inputs, is available at both 2B and 7B scales, and is directly compatible with the VLM2Vec initialization considered in our experiments. Its shared multimodal sequence interface allows instructions, visual tokens, bounding boxes, and textual geo-coordinates to be processed without introducing separate modality-specific encoders. The choice was also guided by computational feasibility and checkpoint availability; we do not claim that Qwen2-VL is optimal among all possible VLM backbones.

Table \ref{tab:full_ablation_results} provides a comprehensive ablation of LoRA-based initialization for \textbf{VLM2GeoVec} against the VLM2Vec baselines, comparing models trained \textit{from scratch} (bootstrap from Qwen2-VL, denoted Q) against those initialized from contrastively pre-trained VLM2Vec weights (denoted V). We report the same suite of metrics as in the main evaluation across both 2B and 7B parameter scales. To focus specifically on how ablation affects cross-modal retrieval, we again refer to Table~\ref{tab:retrieval_evaluation}, which shows how each variant performs on the RSITMD, RSICD, and UCM-caption recall metrics.

We also report the mean $\pm$ standard deviation computed over 3 seeds for VLM2GeoVec (default). Across RSMEB, results are overall stable, rankings are unchanged (both against Tables~\ref{tab:retrieval_evaluation} and~\ref{tab:full_ablation_results}), with the standard deviation typically $<1$--$2$ pp.

\begin{table}[t]
\caption{Initialization and model capacity ablations. Methods in columns, datasets/tasks in rows. Rows show top-1 accuracy (\%) for classification datasets, average recall over R@\{1,5,10\} (\%) for cross-modal retrieval tasks, and precision@1 (\%) for other tasks. The Friedman score is the average rank across all tasks (lower is better), and the rank is the position based on that score. Models \textit{trained from scratch} optimize LoRA adapter over base VLM weights (Qwen2-VL, denoted as \textbf{Q}), instead of boostrapping from contrastively-pretrained weights (VLM2Vec, denoted as \textbf{V}). 2B and 7B denote the number of parameters in the base VLM architecture. Uncertainty estimates (last column): mean and standard deviation are computed over three independent runs (different seeds).}
\label{tab:full_ablation_results}
\centering
\small
\resizebox{\linewidth}{!}{
\begin{tabular}{lcccccc|c}
\toprule
  & \multicolumn{2}{c}{VLM2Vec} 
  & \multicolumn{4}{c}{VLM2GeoVec} 
  \\
  \cmidrule(r){2-3}
  \cmidrule(r){4-8}
Dataset               &  2B (\textbf{Q})  &    7B (\textbf{Q}) &    2B (\textbf{Q}) &     2B   (\textbf{V})         & 7B (\textbf{Q})  &      7B  (\textbf{V})   & 7B  (\textbf{V}) (mean $\pm$ std)  \\
\midrule
\rowcolor{Lavender}\multicolumn{8}{l}{\textbf{Classification}} \\
AID                   & 66.85 & 64.25 & 68.50 & \underline{71.00} & 59.00 & \textbf{77.25} & 74.10 $\pm$ 3.13 \\
Million-AID           & 62.37 & 58.92 & 55.70 & \textbf{65.06} &  6.29 & \underline{64.82} & 66.06 $\pm$ 1.10 \\
RSI-CB              & 40.79 & 32.23 &  6.44 & 40.97 & \textbf{46.05} & \underline{44.54} & 46.46 $\pm$ 2.95 \\
EuroSAT               & 38.44 & 21.26 & 34.19 & \underline{42.59} & \textbf{49.04} & 39.89 & 36.21 $\pm$ 3.37 \\
UCM                   & 68.86 & 69.67 & 83.43 & \underline{84.57} & 83.24 & \textbf{90.24} & 89.01 $\pm$ 1.27 \\
PatternNet            & 64.97 & 62.96 & 16.95 & \underline{78.12} & 31.81 & \textbf{79.76} & 78.45 $\pm$ 2.78 \\
\midrule
\rowcolor{Lavender}\multicolumn{8}{l}{\textbf{Retrieval}} \\
RSITMD I2T            & 27.06 & \underline{29.27} & 25.59 & 28.47 & 29.05 & \textbf{31.86} & 32.13 $\pm$ 0.88 \\
RSITMD T2I            & 31.24 & 34.62 & 33.57 & 35.21 & \underline{37.49} & \textbf{41.03} & 40.06 $\pm$ 0.85 \\
RSICD I2T             & 20.16 & \textbf{23.64} & 19.88 & 19.89 & 21.32 & \underline{21.38} & 22.15 $\pm$ 0.88 \\
RSICD T2I             & 17.75 & 25.90 & 23.88 & 24.07 & \textbf{30.17} & \underline{28.31} & 27.88 $\pm$ 0.49 \\
UCM-caption I2T       & 43.97 & 42.86 & 43.90 & 45.24 & \underline{45.43} & \textbf{48.09} & 51.11 $\pm$ 0.73 \\
UCM-caption T2I       & 45.49 & 49.59 & 51.24 & 48.95 & \textbf{55.14} & \underline{52.76} & 54.16 $\pm$ 0.71 \\
rCIR                  &  2.75 &  1.98 &  9.57 &  9.08 & \underline{19.53} & \textbf{22.99} & 22.99 $\pm$ 0.99 \\
\midrule
\rowcolor{Lavender}\multicolumn{8}{l}{\textbf{Visual Grounding}} \\
RefExp                & 4.10  &  6.65 & 18.75 & 21.55 & \underline{27.75} & \textbf{32.50} & 32.33 $\pm$ 0.14 \\
\midrule
\rowcolor{Lavender}\multicolumn{8}{l}{\textbf{Spatial Localization}} \\
RegCap                &  1.18 &  1.25 & 21.38 & 18.81 & \textbf{27.99} & \underline{26.56} & 25.24 $\pm$ 1.38 \\
GrT2I                 &  0.49 &  0.86 &  2.87 &  4.83 & \underline{10.46} & \textbf{13.70} & 13.74 $\pm$ 0.37 \\
\midrule
\rowcolor{Lavender}\multicolumn{8}{l}{\textbf{Semantic Reverse Geo-localization}} \\
GeoT2I                &  2.80 &  4.40 &  8.05 & 10.60 & \underline{15.25} & \textbf{17.80} & 17.23 $\pm$ 0.90 \\
\midrule
\rowcolor{Lavender}\multicolumn{8}{l}{\textbf{VQA}} \\
LRBEN Presence        & 47.11 & 62.03 & 75.30 & 84.06 & \underline{87.45} & \textbf{89.78} & 89.49 $\pm$ 0.25 \\
LRBEN Comparison      & 65.77 & 38.40 & 62.09 & 79.06 & \underline{88.11} & \textbf{90.33} & 90.30 $\pm$ 0.12 \\
LRBEN Rural–Urban     & 64.00 & 44.00 & 57.00 & 76.00 & \underline{79.00} & \textbf{86.00} & 85.00 $\pm$ 1.73 \\
HRBEN Presence        & 46.95 & 59.12 & 39.67 & 50.20 & \underline{65.43} & \textbf{69.47} & 68.07 $\pm$ 2.05 \\
HRBEN Comparison      & 65.06 & 43.11 & 55.94 & 69.45 & \underline{77.97} & \textbf{79.81} & 79.35 $\pm$ 0.79 \\
\midrule
\rowcolor{LightCyan} 
Score & 4.86 & 4.59 & 4.59 & 3.14 & 2.45 & \textbf{1.36} & \\
\rowcolor{LightCyan}
Rank  & 6      & 4      & 4      & 3      & 2      & \textbf{1} &      \\
\bottomrule

\end{tabular}
}
\end{table}

\subsection{Instruction-Tuning and Robustness}
\label{sec:inst_ablation}
Inspired by recent work, we leverage a diverse instruction pool to prevent overfitting to any single prompt. Our strategy aims to strike a balance between performance and robustness. To investigate the instruction-tuning strategy, we evaluate \textbf{four training-inference settings}:
\begin{enumerate}
    \item \textbf{Diverse-known:} train with 10 varied templates per task; evaluate on a \emph{known} template.
    \item \textbf{Single-known:} train with a single template; evaluate on that \emph{same known} template.
    \item \textbf{Diverse-unseen:} train with 10 varied templates per task; evaluate on an \emph{unseen, re-phrased} template.
    \item \textbf{Single-unseen:} train with a single template; evaluate on an \emph{unseen, re-phrased} template.
\end{enumerate}
Due to time and resource constraints, we use the smaller \textbf{VLM2GeoVec-2B} model variant for these ablations. Results for all settings are summarized in Table~\ref{tab:inst_ablation}.

\textbf{Impact of diverse instructions.}
When we evaluate on known prompts, \textbf{diverse-known} and \textbf{single-known} are very close overall. The classification benchmarks (AID, Million-AID, RSI-CB, EuroSAT, UCMerced, PatternNet) slightly prefer diverse-known (4/6 wins), while the VQA benchmarks favor single-known (4/5 wins). This indicates that template diversity does not consistently improve peak in-distribution accuracy---especially for VQA---but it also does not introduce noticeable degradation.

\textbf{Robustness in the presence of unseen phrasings.}
To evaluate generalization to novel prompts with re-phrased wording, we compared performance changes when moving from known to unseen instruction sets under both diverse- and single-instruction training. Across RSMEB, the diverse-instruction strategy is substantially more stable, with an average drop of \textbf{3.53 pp}, whereas the single-instruction strategy suffers a much larger \textbf{10.05 pp} drop. The largest robustness gaps remain on PatternNet (\textbf{49.70 pp vs.\ 60.52 pp}) and RSI-CB (\textbf{29.24 pp vs.\ 40.23 pp}), highlighting where unseen prompts are most damaging. Notably, a few VQA tasks under diverse training even improve with unseen templates, suggesting that prompt rephrasing can sometimes help but is far less predictable under single-instruction training.

Consistent with this trend, the Friedman average ranks over all metrics place \textbf{diverse-known first (2.16)}, ahead of single-known (2.27), diverse-unseen (2.34), and single-unseen (3.23). Overall, these experiments demonstrate that while template diversity may not always boost peak in-distribution accuracy, it plays a crucial role in ensuring robustness to new or unexpected phrasing. Future work could therefore focus on methods for selecting or weighting templates to strike the optimal balance between in-distribution performance and out-of-distribution generalization.

\begin{table}[t]
\caption{Instruction-tuning and robustness ablations. Methods in columns, datasets/tasks in rows. Rows show top-1 accuracy (\%) for classification datasets, average recall over R@\{1,5,10\} (\%) for cross-modal retrieval tasks, and precision@1 (\%) for other tasks. Friedman score is the average rank across all tasks (lower is better), and rank is the position based on that score.}
\centering
\small
\resizebox{\linewidth}{!}{
\begin{tabular}{lcccc}
\toprule
\textbf{Dataset} & \textbf{diverse-known} & \textbf{single-known} & \textbf{diverse-unseen} & \textbf{single-unseen} \\
\midrule
\rowcolor{Lavender}\multicolumn{5}{l}{\textbf{Classification}} \\
AID                  & 71.00 & 66.30 & 70.85 & 72.75 \\
Million-AID          & 65.06 & 63.16 & 57.11 & 58.33 \\
RSI-CB               & 40.97 & 47.25 & 11.73 & 7.02 \\
EuroSAT              & 42.59 & 44.78 & 41.37 & 45.74 \\
UCMerced             & 84.57 & 82.43 & 85.00 & 81.67 \\
PatternNet           & 78.12 & 76.36 & 28.42 & 15.84 \\
\midrule
\rowcolor{Lavender}\multicolumn{5}{l}{\textbf{Retrieval}} \\
RSITMD I2T           & 28.47 & 30.24 & 29.72 & 31.86 \\
RSITMD T2I           & 35.21 & 32.86 & 34.87 & 32.58 \\
RSICD I2T            & 19.89 & 21.26 & 23.15 & 22.63 \\
RSICD T2I            & 24.07 & 23.45 & 24.74 & 22.32 \\
UCM-caption I2T      & 45.24 & 41.59 & 45.08 & 43.49 \\
UCM-caption T2I      & 48.95 & 45.46 & 51.27 & 47.15 \\
rCIR                 & 9.08  & 12.76 & 8.80  & 8.42 \\
\midrule
\rowcolor{Lavender}\multicolumn{5}{l}{\textbf{Visual Grounding}} \\
RefExp               & 21.55 & 20.45 & 17.40 & 15.85 \\
\midrule
\rowcolor{Lavender}\multicolumn{5}{l}{\textbf{Spatial Localization}} \\
RegCap               & 18.81 & 20.09 & 6.65  & 7.94 \\
GrT2I                & 4.83  & 3.06  & 4.83  & 2.94 \\
\midrule
\rowcolor{Lavender}\multicolumn{5}{l}{\textbf{Semantic Reverse Geo-localization}} \\
GeoT2I               & 10.60 & 10.95 & 9.65  & 5.45 \\
\midrule
\rowcolor{Lavender}\multicolumn{5}{l}{\textbf{VQA}} \\
LRBEN Presence       & 84.06 & 89.07 & 77.53 & 27.17 \\
LRBEN Comparison     & 79.06 & 88.33 & 80.18 & 66.89 \\
LRBEN Rural-Urban    & 73.00 & 67.00 & 84.00 & 60.00 \\
HRBEN Presence       & 50.20 & 61.21 & 62.88 & 61.22 \\
HRBEN Comparison     & 69.45 & 77.10 & 71.86 & 66.75 \\
\midrule
\rowcolor{LightCyan} 
Score & \textbf{2.16} & 2.27 & 2.34 & 3.23 \\
\rowcolor{LightCyan} 
Rank                 & \textbf{1}    & 2    & 3    & 4    \\
\bottomrule
\end{tabular}
}

\label{tab:inst_ablation}
\end{table}

\clearpage

\section{Qualitative Results}
\label{sec:qualitative_results_extra}

\subsection{Success Cases}

We illustrate the typical behavior of the model on three retrieval tasks using border color conventions throughout the figures:
\textcolor{blue}{\textbf{blue}} for query images,  
\textcolor{green}{\textbf{green}} for correct targets and
\textcolor{red}{\textbf{red}} for the remaining top-5 candidates.  
Query prompts (instruction and input text) appear above each image gallery.

\begin{figure}[t]
  \centering
  \includegraphics[width=0.8\textwidth,
                   keepaspectratio]{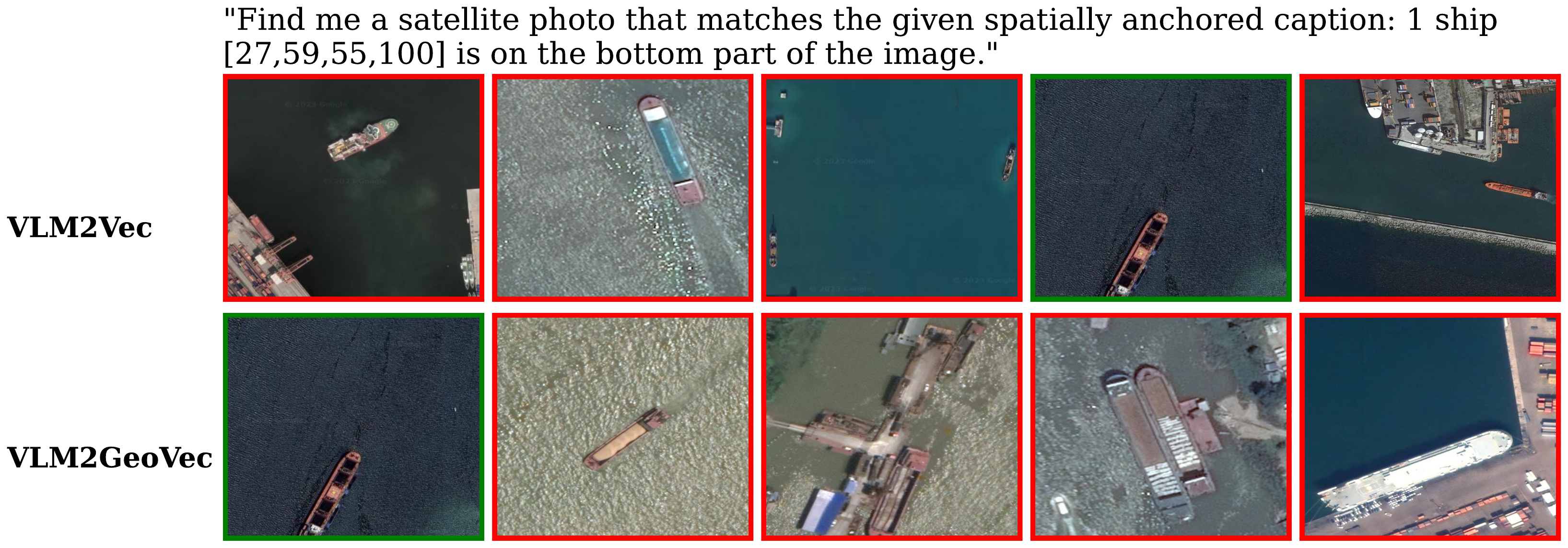}
  \vfill                    
  \includegraphics[width=0.8\textwidth,
                   keepaspectratio]{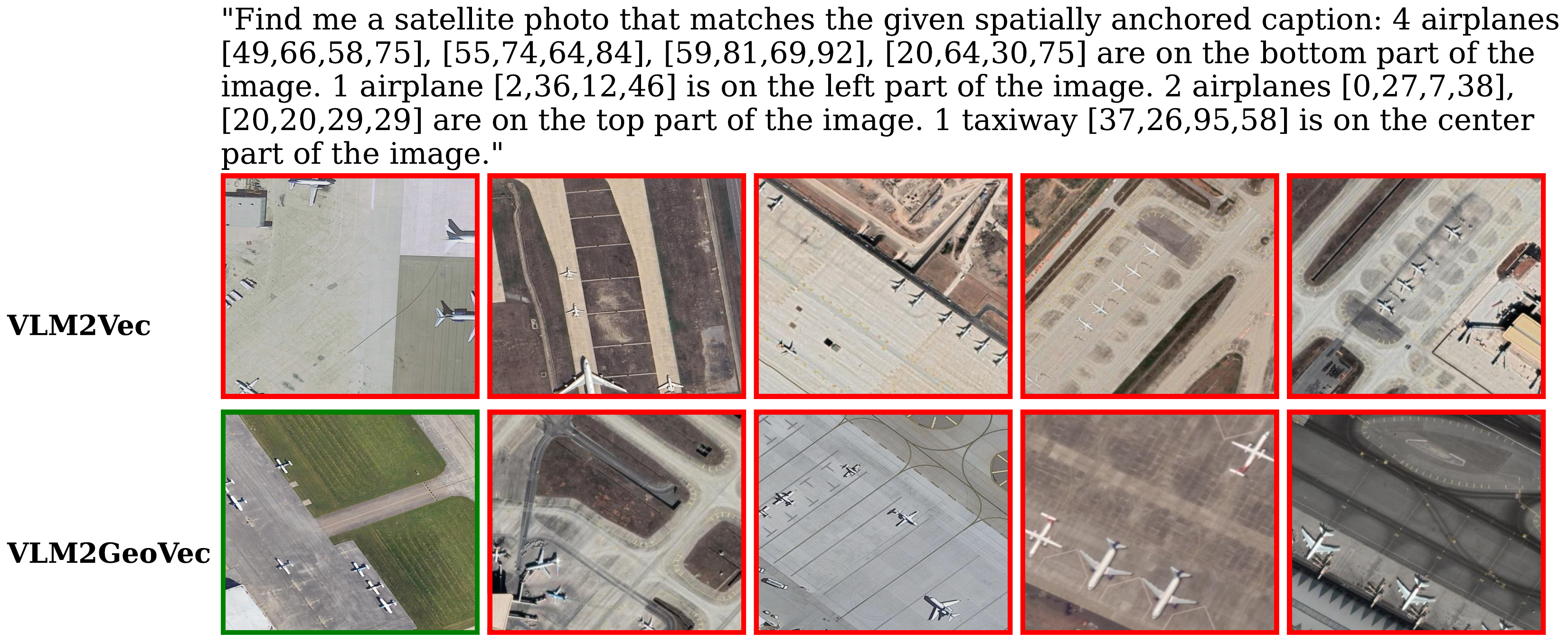}
  \vfill 
  \includegraphics[width=0.8\textwidth,
                   keepaspectratio]{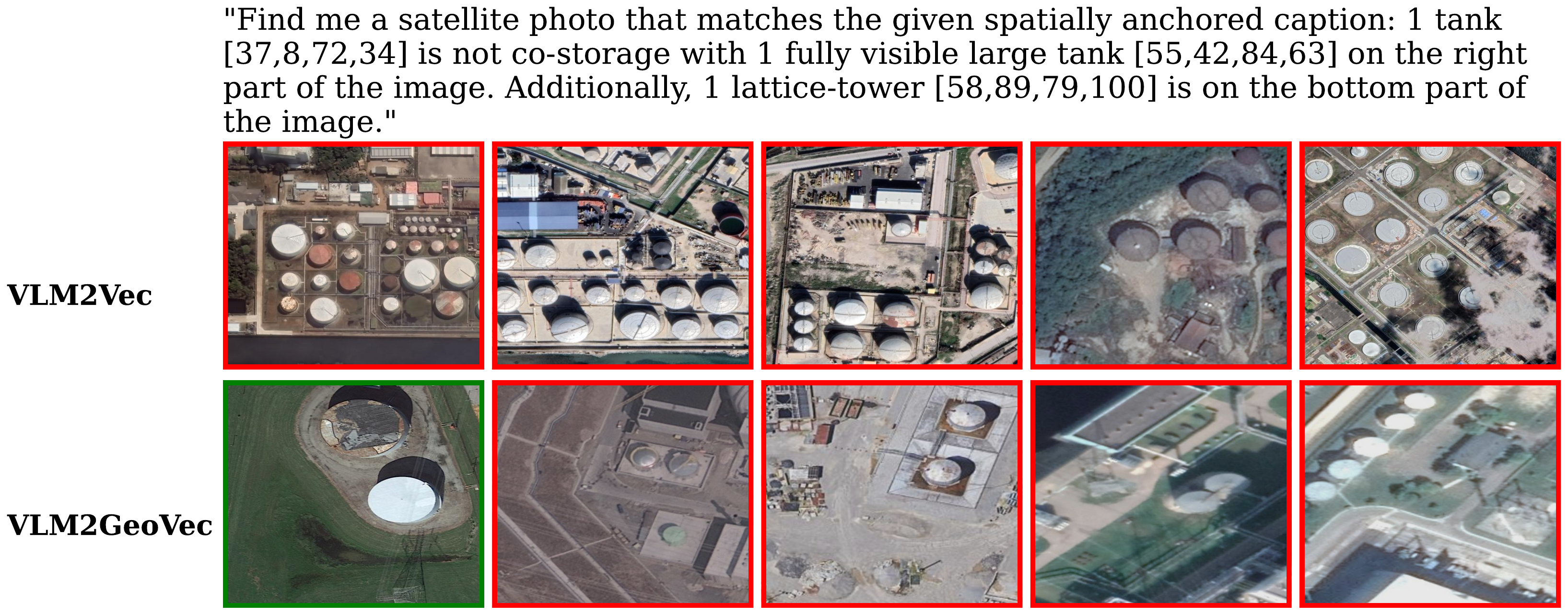}
  \vfill 
  \includegraphics[width=0.8\textwidth,
                   keepaspectratio]{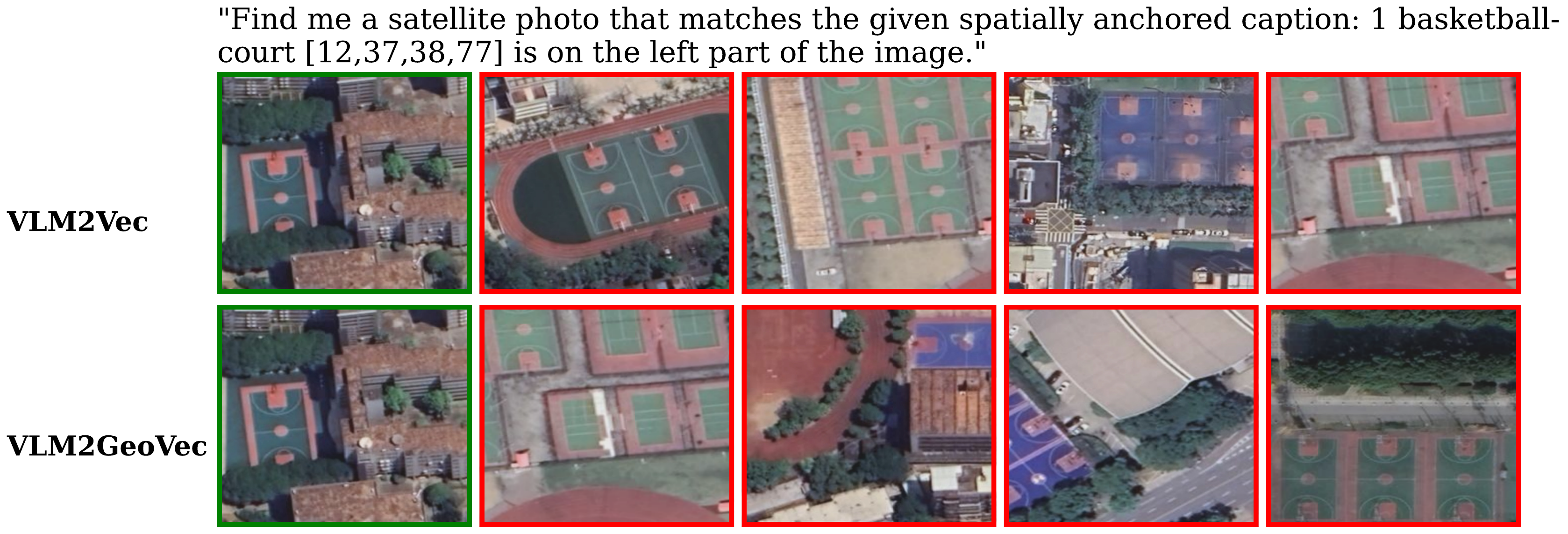}
  \caption{Examples of top-5 retrieved candidates for grounded text-to-image retrieval (GrT2I). Each panel shows an instruction and the text query with bounding boxes. The ground-truth image appears in \textcolor{green}{green}, while the other four retrieved images are in \textcolor{red}{red}.}
  \label{fig:grounded-t2i-samples}
\end{figure}

\begin{figure}[t]
  \centering
  \includegraphics[width=0.8\textwidth,
                   keepaspectratio]{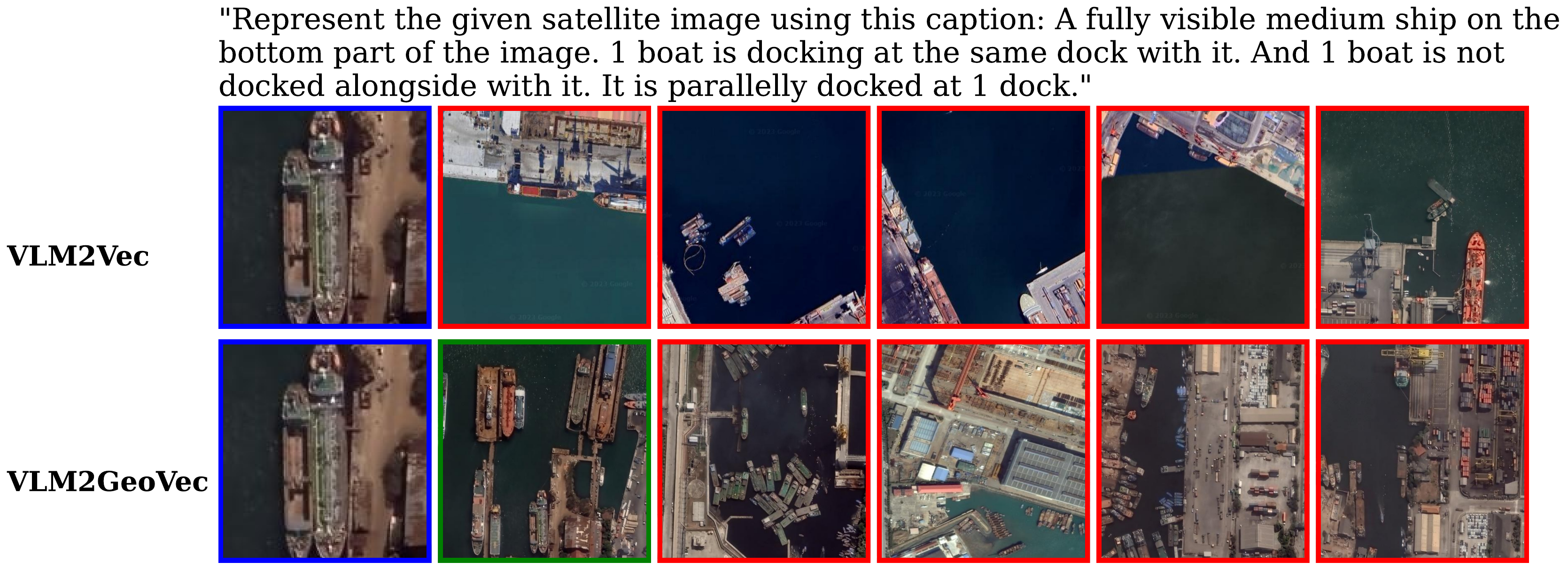}
  \vfill                     
  \includegraphics[width=0.8\textwidth,
                   keepaspectratio]{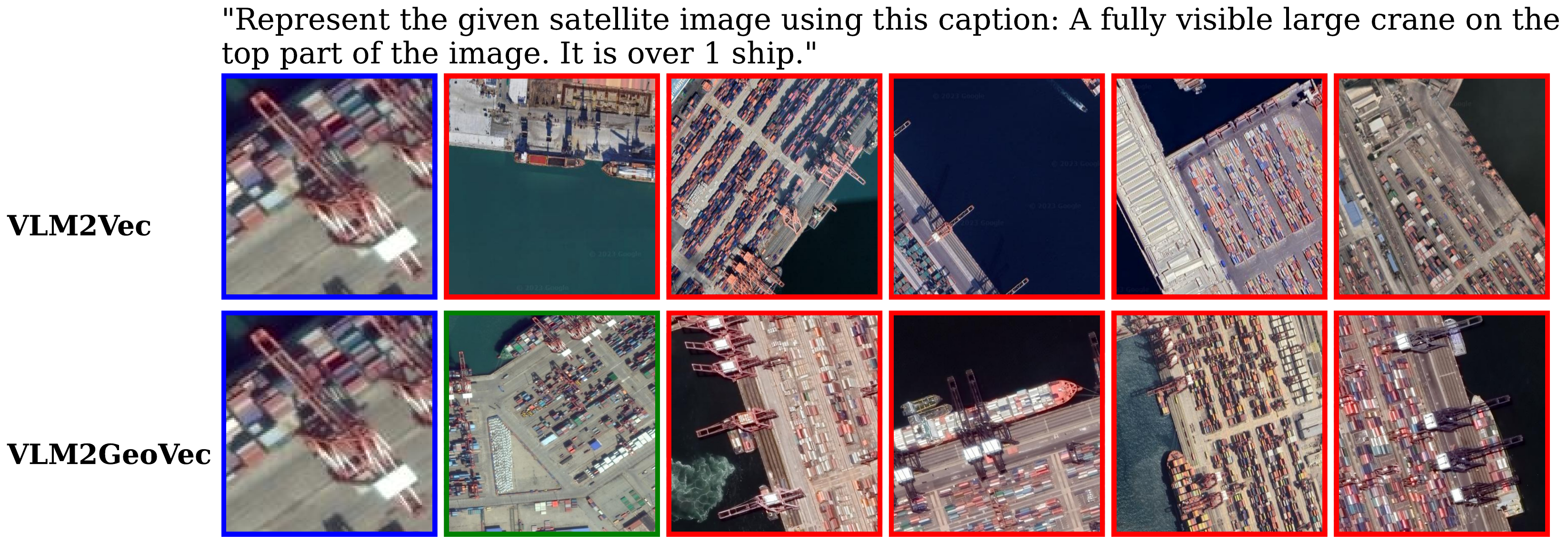}
  \vfill                     
  \includegraphics[width=0.8\textwidth,
                   keepaspectratio]{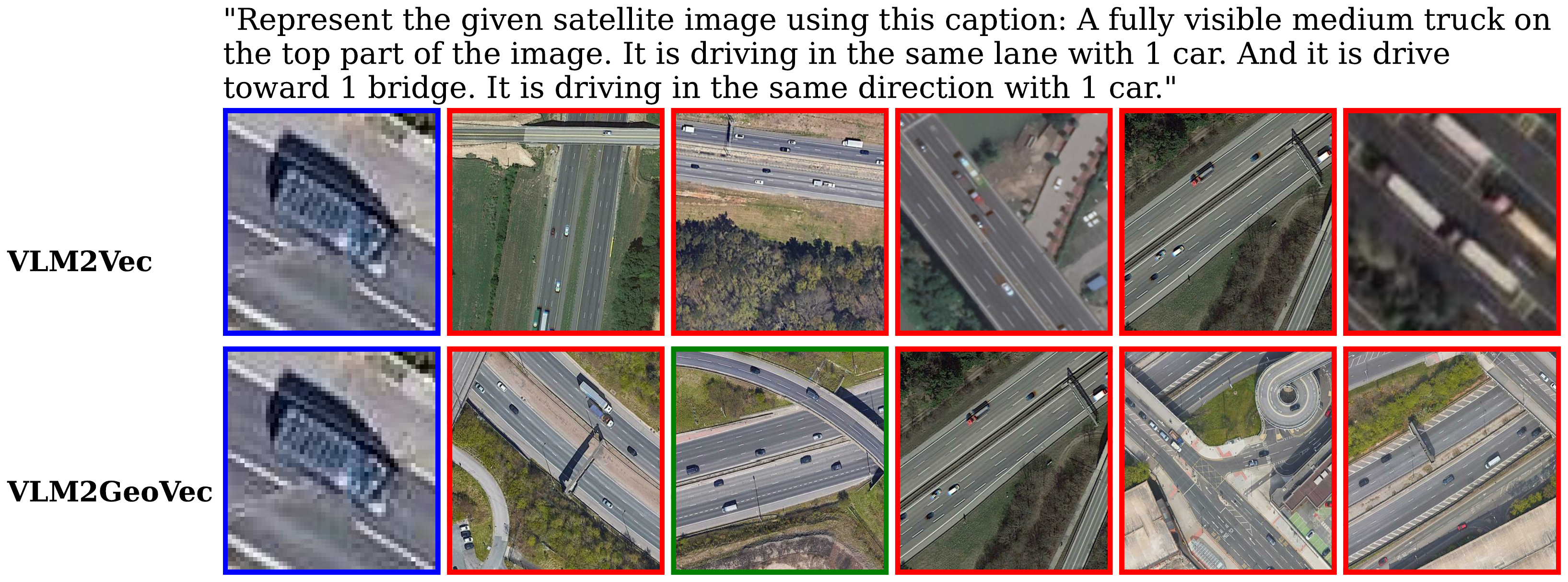}
  \vfill                     
  \includegraphics[width=0.8\textwidth,
                   keepaspectratio]{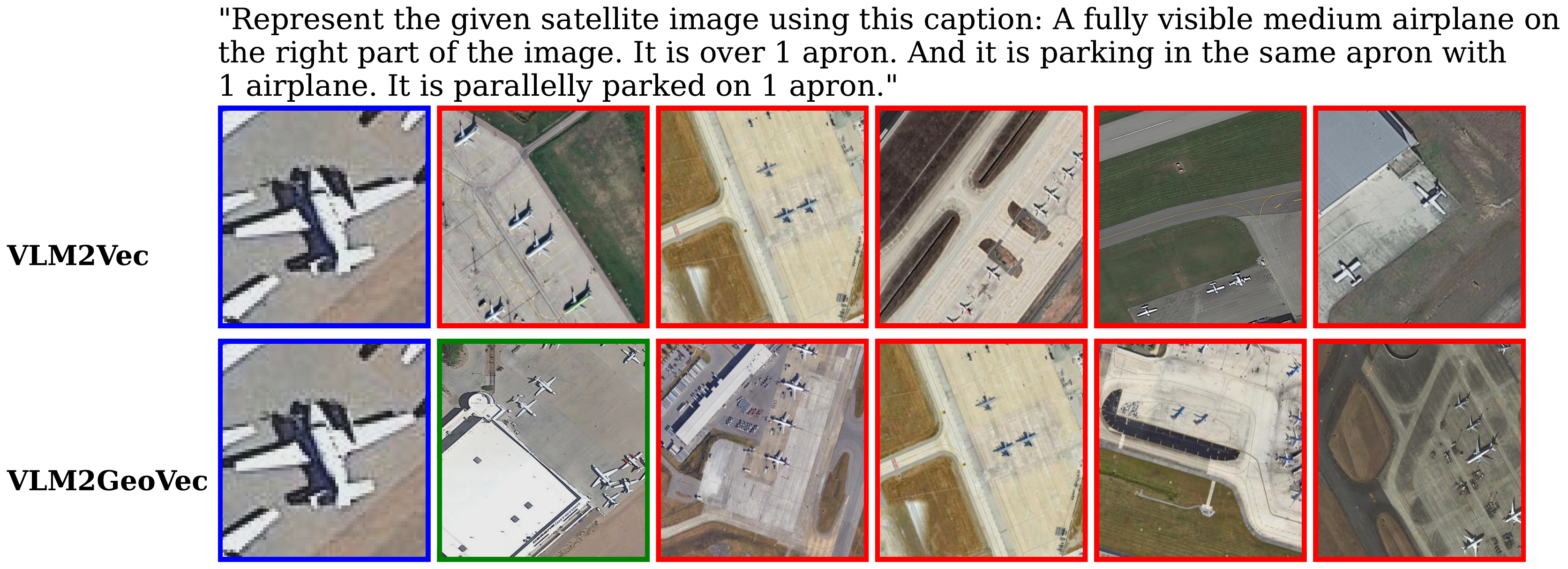}
  \caption{Examples of top-5 retrieved candidates for region-based composed image retrieval (rCIR). The input query consists of an image region plus an edit instruction. The query region is outlined in \textcolor{blue}{blue}, the correctly modified target region in \textcolor{green}{green}, and the other candidates in \textcolor{red}{red}, illustrating how well the model applies the specified change.}
  \label{fig:rcir-samples}
\end{figure}

\begin{figure}[t]
  \centering
  \includegraphics[width=0.8\textwidth,
                   keepaspectratio]{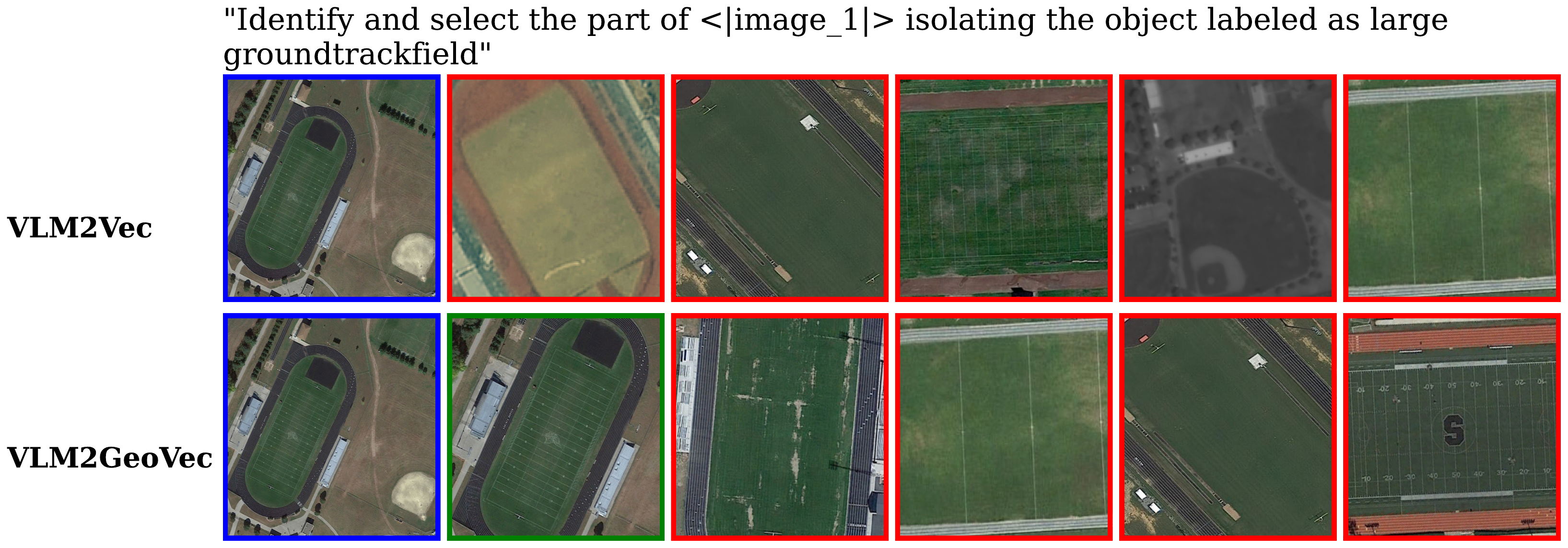}
  \vfill                     
  \includegraphics[width=0.8\textwidth,
                   keepaspectratio]{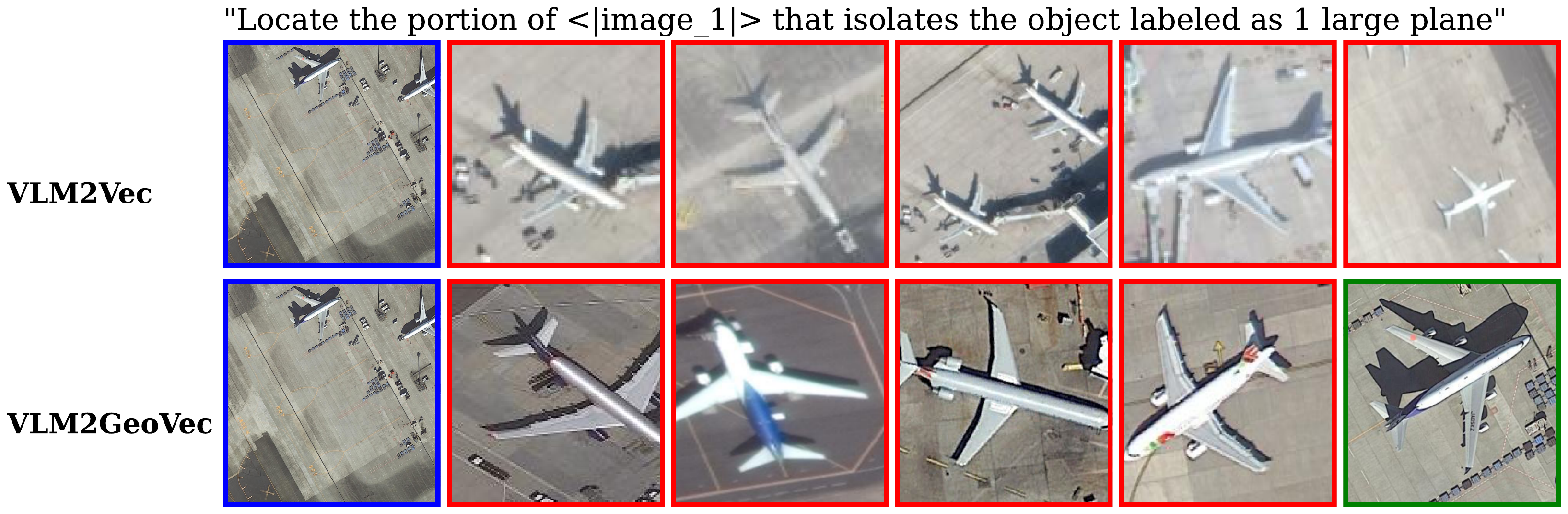}
  \vfill                     
  \includegraphics[width=0.8\textwidth,
                   keepaspectratio]{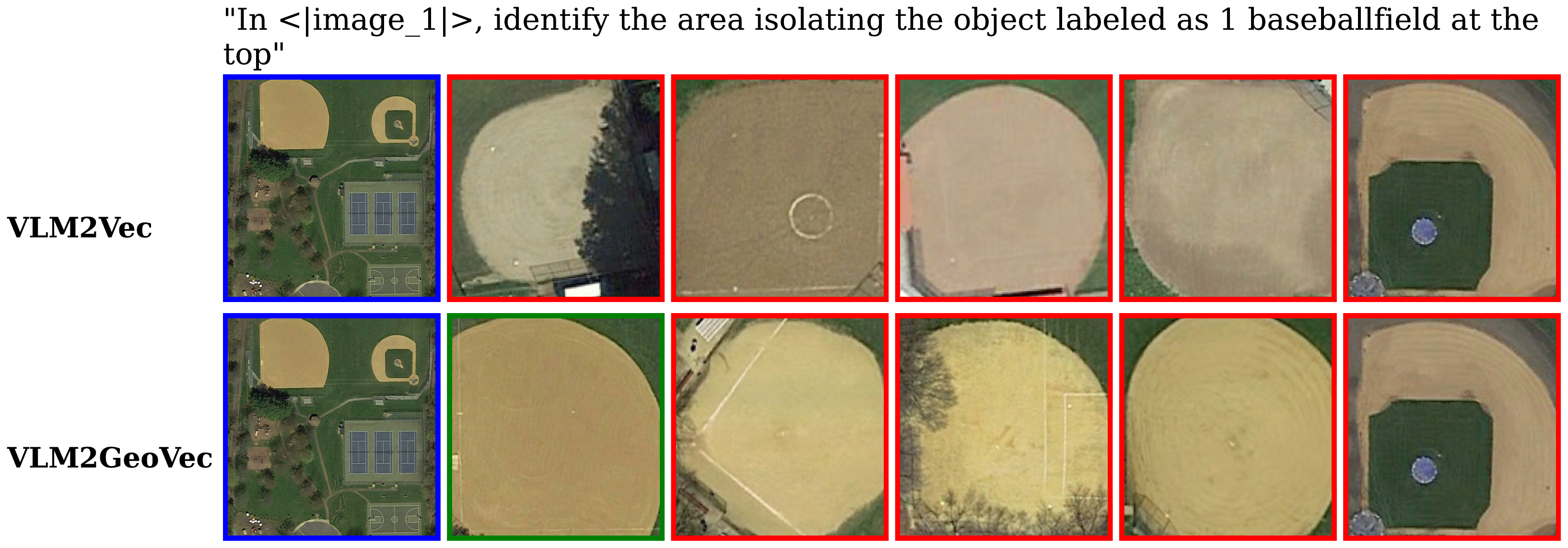}
  \vfill                     
  \includegraphics[width=0.8\textwidth,
                   keepaspectratio]{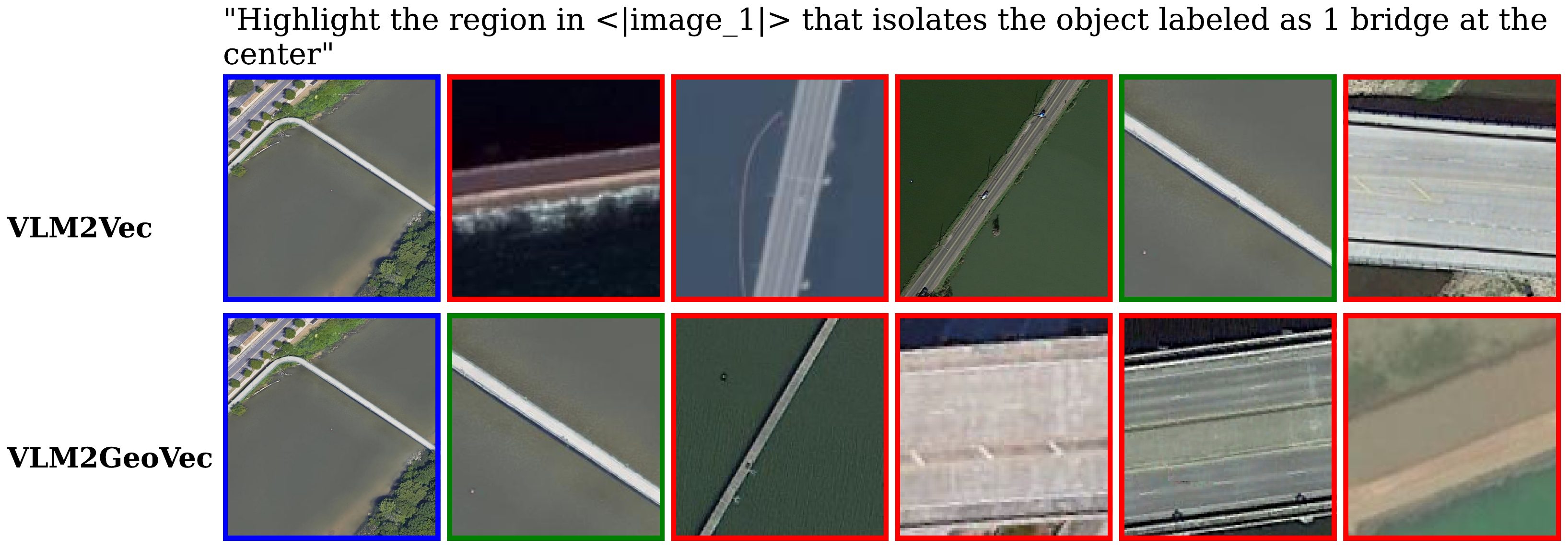}
  \caption{Examples of top-5 retrieved candidates for referring-expression retrieval (RefExp). For each example, the full image is outlined in \textcolor{blue}{blue} and paired with a referring expression. The model’s correct region selection is shown in \textcolor{green}{green}, and the alternative proposals in \textcolor{red}{red}, highlighting its localization accuracy.}
  \label{fig:refexp-samples}
\end{figure}

\begin{itemize}
  \item \textbf{Figure \ref{fig:grounded-t2i-samples}} (Grounded Text-to-Image Retrieval):  
        Each panel shows an instruction and the text query with bounding boxes. The ground-truth image appears in \textcolor{green}{green}, while the other four retrieved images are in \textcolor{red}{red}.

  \item \textbf{Figure \ref{fig:rcir-samples}} (Region-based Composed Image Retrieval):  
        The input query consists of an image region plus an edit instruction. The query region is outlined in \textcolor{blue}{blue}, the correctly modified target region in \textcolor{green}{green}, and the other candidates in \textcolor{red}{red}, illustrating how well the model applies the specified change.

  \item \textbf{Figure \ref{fig:refexp-samples}} (Referring-Expression Retrieval):  
        For each example, the full image is outlined in \textcolor{blue}{blue} and paired with a referring expression. The model’s correct region selection is shown in \textcolor{green}{green}, and the alternative proposals in \textcolor{red}{red}, highlighting its localization accuracy.
\end{itemize}

\clearpage

\clearpage

\clearpage

\clearpage

\subsection{Failure Cases}

\begin{figure}[t]
  \centering
  \includegraphics[width=0.8\textwidth,
                   keepaspectratio]{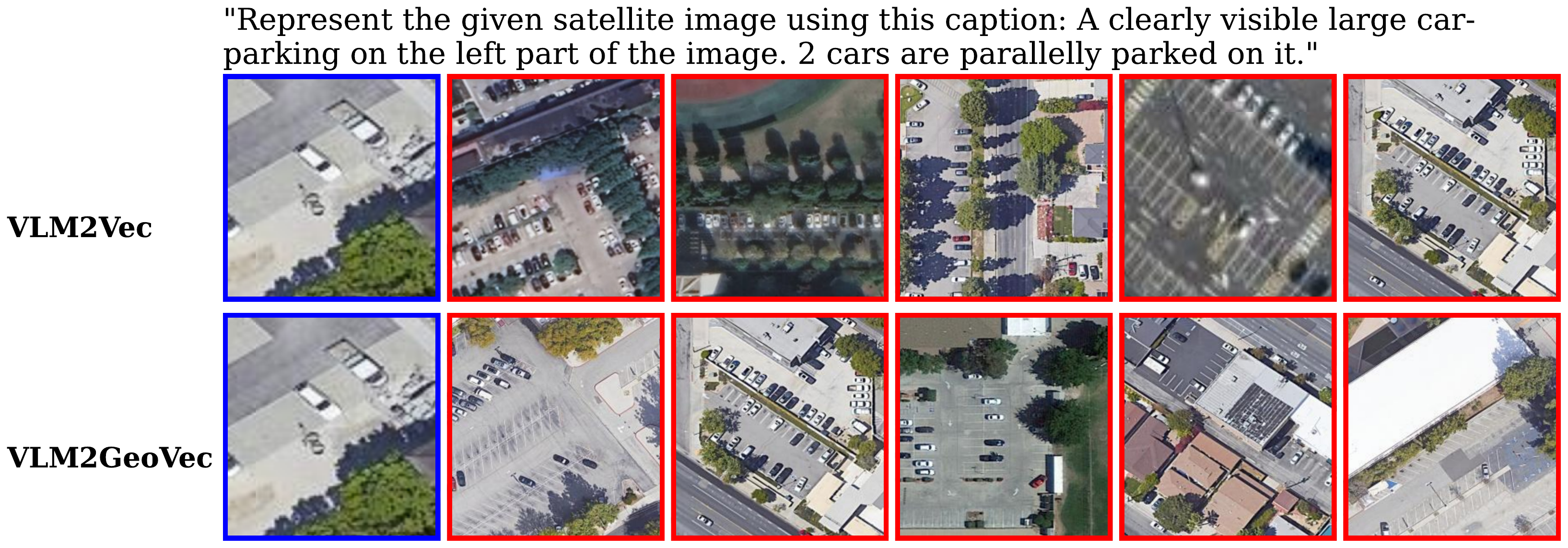}
  \vfill                     
  \includegraphics[width=0.8\textwidth,
                   keepaspectratio]{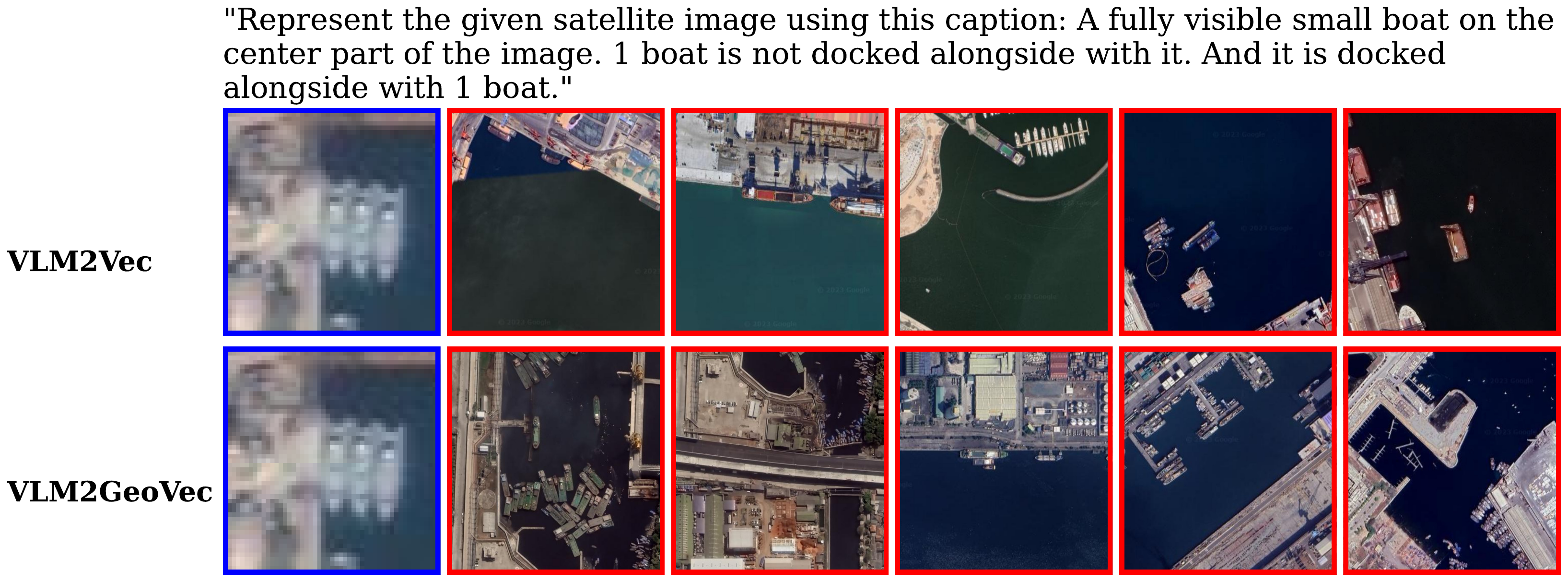}
  \caption{Failure examples for region-based composed image retrieval (rCIR). The input query consists of an image region plus an edit instruction. The query region is outlined in \textcolor{blue}{blue}; incorrect candidates are highlighted in \textcolor{red}{red}.}
  \label{fig:rcir-failures}
\end{figure}

\begin{figure}[t]
  \centering
  \includegraphics[width=0.8\textwidth,
                   keepaspectratio]{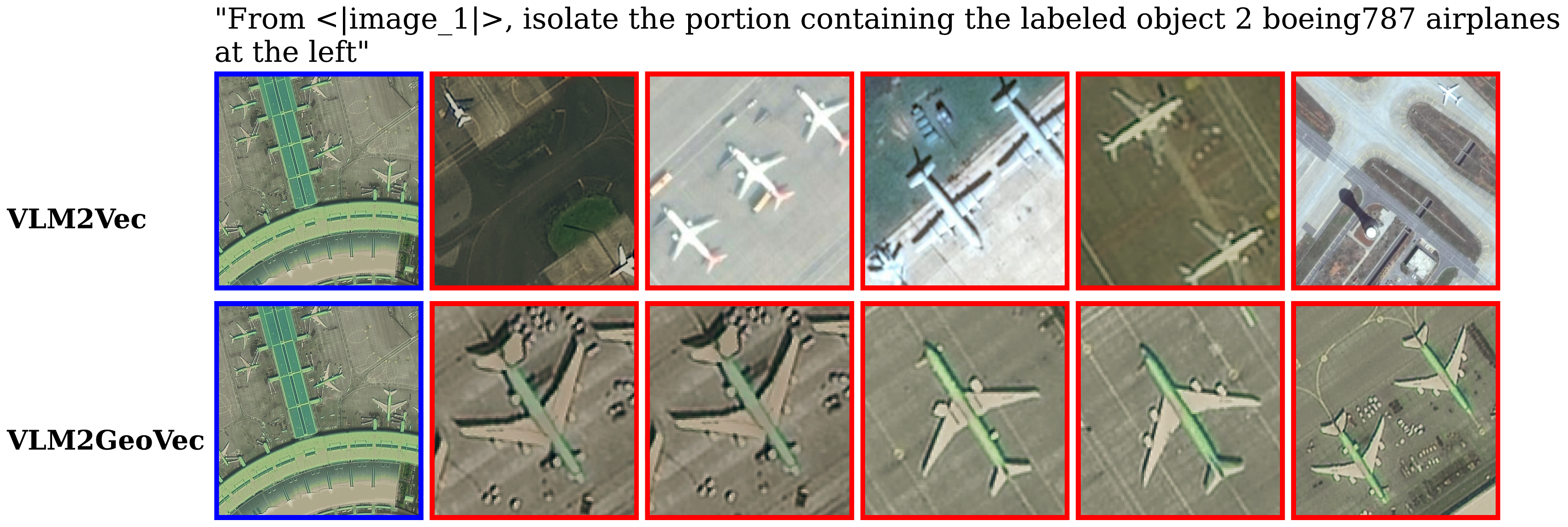}
  \vfill                     
  \includegraphics[width=0.8\textwidth,
                   keepaspectratio]{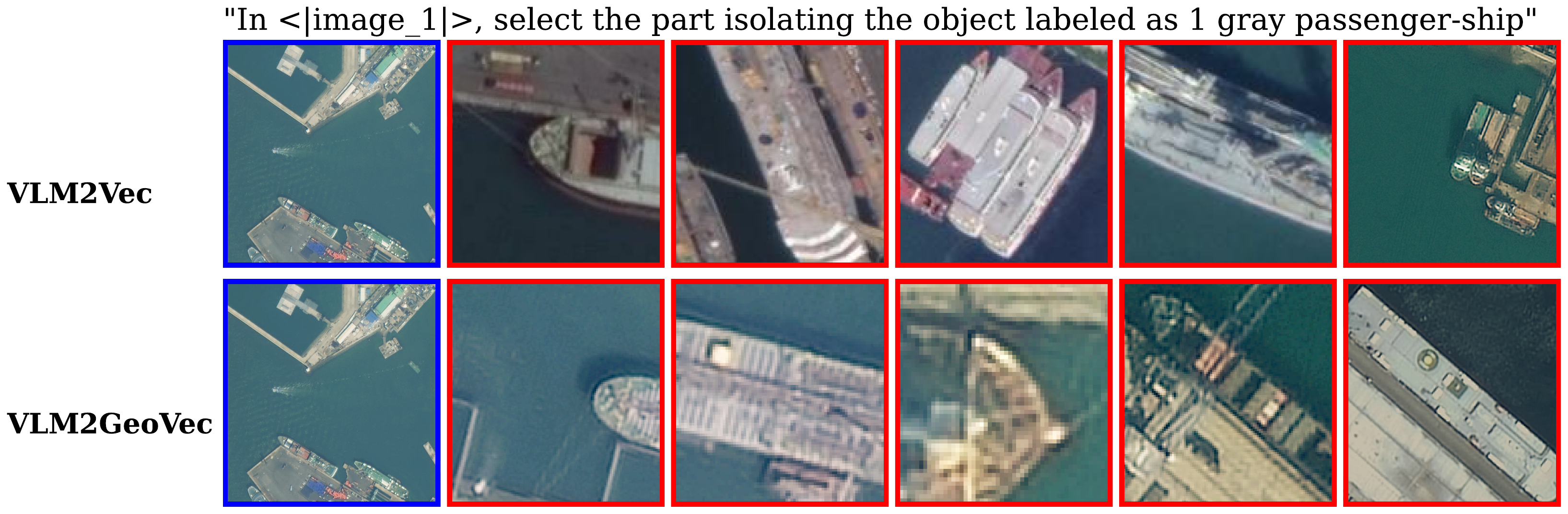}
  \caption{Failure examples for referring-expression retrieval (RefExp). For each example, the full image is outlined in \textcolor{blue}{blue} and accompanied by a corresponding referring expression; incorrect candidates are highlighted in \textcolor{red}{red}.}
  \label{fig:refexp-failures}
\end{figure}

We also show examples when the model fails to retrieve the correct target within the top-5 candidates:

\begin{itemize}

  \item \textbf{Figure \ref{fig:rcir-failures}} (Region-based Composed Image Retrieval):  
        The input query consists of an image region plus an edit instruction. The query region is outlined in \textcolor{blue}{blue}, and the incorrectly retrieved candidates are in \textcolor{red}{red}. While VLM2GeoVec retrieves images with a similar background (e.g., parking lot) and type of objects, it has issues reasoning about the number of objects.

  \item \textbf{Figure \ref{fig:refexp-failures}} (Referring-Expression Retrieval):  
        For each example, the full image is outlined in \textcolor{blue}{blue} and accompanied by a corresponding referring expression, and the incorrectly retrieved candidates are in \textcolor{red}{red}. We observe that VLM2GeoVec retrieves images with the correct object type and style (e.g., green planes), but it can struggle with the correct combination of object orientation and count.
\end{itemize}

\clearpage

These failures indicate that the model often retrieves candidates with the correct semantic category and visual context but remains sensitive to object count, orientation, and relational composition. This suggests that the unified embedding captures broad scene semantics more reliably than precise structured relations.

\clearpage

\section{Task Prompts}
\label{sec:instructions_extra}

For instruction-following embedding models, we use the following set of query prompts to guide each task during inference:

\begin{itemize}
  \item \textbf{Classification (VLM2Vec):}\newline
    \textit{<|image\_pad|> Represent the given image for classification.}

  \item \textbf{Classification (VLM2GeoVec):}\newline
    \textit{<|image\_pad|> Find an image caption describing the given satellite image.}

  \item \textbf{Text-to-image retrieval:}\newline
    \textit{Find me a satellite image that matches the given caption: ...}

  \item \textbf{Image-to-text retrieval:}\newline
    \textit{<|image\_pad|> Find an image caption describing the given satellite image.}

  \item \textbf{Region-based composed image retrieval:}\newline
    \textit{<|image\_pad|> Represent the given satellite image using this caption: ...}

  \item \textbf{VQA:}\newline
    \textit{<|image\_pad|> Represent the given image with the following question: ...}

  \item \textbf{Referring-expression retrieval:}\newline
    \textit{<|image\_pad|> Select the portion of the satellite image that isolates the object labeled as ...}

  \item \textbf{Region-caption retrieval:}\newline
    \textit{<|image\_pad|> Identify the object shown in the image within the region ...}

  \item \textbf{Grounded text-to-image retrieval:}\newline
    \textit{Find me a satellite photo that matches the given spatially anchored caption: ...}

  \item \textbf{Geo-localized text-to-image retrieval:}\newline
    \textit{Find me a satellite image that matches the given caption at <|latitude\_longitude|>: ...}
\end{itemize}

We denote as <|image\_pad|> and <|latitude\_longitude|> the sequence of image tokens and textual geo-coordinates, respectively. For target prompts, we use the following set of instructions:

\begin{itemize}
  \item \textbf{Most tasks with a target image:}\newline
    \textit{<|image\_pad|> Represent the given image.}

  \item \textbf{Text-to-image retrieval:}\newline
    \textit{<|image\_pad|> Find an image caption describing the given satellite image.}    

  \item \textbf{Referring-expression retrieval:}\newline
    \textit{<|image\_pad|> Represent the given cropped image of the object.}
\end{itemize}

For \textbf{classification} tasks, we follow CLIP’s prompt‐ensembling: for each class label, we instantiate all 20 prefixes below, encode each resulting prompt, average their embeddings, and compare to the image embedding via cosine similarity.

\begin{itemize}
  \item \textit{satellite imagery of [class label]}
  \item \textit{aerial imagery of [class label]}
  \item \textit{a satellite photo of [class label]}
  \item \textit{an aerial photo of [class label]}
  \item \textit{a satellite view of [class label]}
  \item \textit{an aerial view of [class label]}
  \item \textit{satellite imagery of a [class label]}
  \item \textit{aerial imagery of a [class label]}
  \item \textit{a satellite photo of a [class label]}
  \item \textit{an aerial photo of a [class label]}
  \item \textit{a satellite view of a [class label]}
  \item \textit{an aerial view of a [class label]}
  \item \textit{satellite imagery of the [class label]}
  \item \textit{aerial imagery of the [class label]}
  \item \textit{a satellite photo of the [class label]}
  \item \textit{an aerial photo of the [class label]}
  \item \textit{a satellite view of the [class label]}
  \item \textit{an aerial view of the [class label]}
  \item \textit{a satellite image of [class label]}
  \item \textit{an aerial image of [class label]}
\end{itemize}

%% file: main.bib
@String(ICCV  = {Int. Conf. Comput. Vis.})

@String(ECCV  = {Eur. Conf. Comput. Vis.})

@String(AAAI  = {Nat. Conf. on Art. Int.})

@String(ICCV  = {ICCV})

@String(ECCV  = {ECCV})

@String(AAAI  = {AAAI})

@inproceedings{kuckreja2024geochat,
  title={Geochat: Grounded large vision-language model for remote sensing},
  author={Kuckreja, Kartik and Danish, Muhammad Sohail and Naseer, Muzammal and Das, Abhijit and Khan, Salman and Khan, Fahad Shahbaz},
  booktitle={Proceedings of the IEEE/CVF Conference on Computer Vision and Pattern Recognition},
  pages={27831--27840},
  year={2024}
}

@inproceedings{irvin2024teochat,
  title={TEOChat: A Large Vision-Language Assistant for Temporal Earth Observation Data},
  author={Irvin, Jeremy Andrew and Liu, Emily Ruoyu and Chen, Joyce Chuyi and Dormoy, Ines and Kim, Jinyoung and Khanna, Samar and Zheng, Zhuo and Ermon, Stefano},
  booktitle={International Conference on Learning Representations},
  year={2025}
}

@inproceedings{radford2021clip,
  title={Learning transferable visual models from natural language supervision},
  author={Radford, Alec and Kim, Jong Wook and Hallacy, Chris and Ramesh, Aditya and Goh, Gabriel and Agarwal, Sandhini and Sastry, Girish and Askell, Amanda and Mishkin, Pamela and Clark, Jack and others},
  booktitle={International conference on machine learning},
  pages={8748--8763},
  year={2021},
  organization={PmLR}
}

@inproceedings{Zhai2023SigLIP,
  title={Sigmoid loss for language image pre-training},
  author={Zhai, Xiaohua and Mustafa, Basil and Kolesnikov, Alexander and Beyer, Lucas},
  booktitle={Proceedings of the IEEE/CVF international conference on computer vision},
  pages={11975--11986},
  year={2023}
}

@inproceedings{
jiang2024vlm2vec,
title={{VLM}2Vec: Training Vision-Language Models for Massive Multimodal Embedding Tasks},
author={Ziyan Jiang and Rui Meng and Xinyi Yang and Semih Yavuz and Yingbo Zhou and Wenhu Chen},
booktitle={The Thirteenth International Conference on Learning Representations},
year={2025},
}

@inproceedings{wei2024uniir,
  title={Uniir: Training and benchmarking universal multimodal information retrievers},
  author={Wei, Cong and Chen, Yang and Chen, Haonan and Hu, Hexiang and Zhang, Ge and Fu, Jie and Ritter, Alan and Chen, Wenhu},
  booktitle={European Conference on Computer Vision},
  pages={387--404},
  year={2024},
  organization={Springer}
}

@inproceedings{
lin2024mmembed,
title={{MM}-{EMBED}: UNIVERSAL MULTIMODAL RETRIEVAL WITH MULTIMODAL {LLMS}},
author={Sheng-Chieh Lin and Chankyu Lee and Mohammad Shoeybi and Jimmy Lin and Bryan Catanzaro and Wei Ping},
booktitle={The Thirteenth International Conference on Learning Representations},
year={2025},
}

@inproceedings{muennighoff2022mteb,
  title={MTEB: Massive Text Embedding Benchmark},
  author={Muennighoff, Niklas and Tazi, Nouamane and Magne, Loic and Reimers, Nils},
  booktitle={Proceedings of the 17th Conference of the European Chapter of the Association for Computational Linguistics},
  pages={2014--2037},
  year={2023}
}

@article{liu2023visual,
  title={Visual instruction tuning},
  author={Liu, Haotian and Li, Chunyuan and Wu, Qingyang and Lee, Yong Jae},
  journal={Advances in neural information processing systems},
  volume={36},
  pages={34892--34916},
  year={2023}
}

@inproceedings{
li2024llava,
title={{LL}a{VA}-Ne{XT}-Interleave: Tackling Multi-image, Video, and 3D in Large Multimodal Models},
author={Feng Li and Renrui Zhang and Hao Zhang and Yuanhan Zhang and Bo Li and Wei Li and Zejun MA and Chunyuan Li},
booktitle={The Thirteenth International Conference on Learning Representations},
year={2025},
}

@misc{liu2024llavanext,
    title={LLaVA-NeXT: Improved reasoning, OCR, and world knowledge},
    url={https://llava-vl.github.io/blog/2024-01-30-llava-next/},
    author={Liu, Haotian and Li, Chunyuan and Li, Yuheng and Li, Bo and Zhang, Yuanhan and Shen, Sheng and Lee, Yong Jae},
    month={January},
    year={2024}
}

@misc{li2024llavanext-strong,
    title={LLaVA-NeXT: Stronger LLMs Supercharge Multimodal Capabilities in the Wild},
    url={https://llava-vl.github.io/blog/2024-05-10-llava-next-stronger-llms/},
    author={Li, Bo and Zhang, Kaichen and Zhang, Hao and Guo, Dong and Zhang, Renrui and Li, Feng and Zhang, Yuanhan and Liu, Ziwei and Li, Chunyuan},
    month={May},
    year={2024}
}

@article{wang2024qwen2,
  title={Qwen2-vl: Enhancing vision-language model's perception of the world at any resolution},
  author={Wang, Peng and Bai, Shuai and Tan, Sinan and Wang, Shijie and Fan, Zhihao and Bai, Jinze and Chen, Keqin and Liu, Xuejing and Wang, Jialin and Ge, Wenbin and others},
  journal={arXiv preprint arXiv:2409.12191},
  year={2024}
}

@InProceedings{Liu_2021CIRR,
    author    = {Liu, Zheyuan and Rodriguez-Opazo, Cristian and Teney, Damien and Gould, Stephen},
    title     = {Image Retrieval on Real-Life Images With Pre-Trained Vision-and-Language Models},
    booktitle = {Proceedings of the IEEE/CVF International Conference on Computer Vision (ICCV)},
    month     = {October},
    year      = {2021},
    pages     = {2125-2134}
}

@inproceedings{jia2021align,
  title={Scaling up visual and vision-language representation learning with noisy text supervision},
  author={Jia, Chao and Yang, Yinfei and Xia, Ye and Chen, Yi-Ting and Parekh, Zarana and Pham, Hieu and Le, Quoc and Sung, Yun-Hsuan and Li, Zhen and Duerig, Tom},
  booktitle={International conference on machine learning},
  pages={4904--4916},
  year={2021},
  organization={PMLR}
}

@inproceedings{li2022blip,
  title={Blip: Bootstrapping language-image pre-training for unified vision-language understanding and generation},
  author={Li, Junnan and Li, Dongxu and Xiong, Caiming and Hoi, Steven},
  booktitle={International conference on machine learning},
  pages={12888--12900},
  year={2022},
  organization={PMLR}
}

@article{li2025benchmark,
  title={Benchmark evaluations, applications, and challenges of large vision language models: A survey},
  author={Li, Zongxia and Wu, Xiyang and Du, Hongyang and Nghiem, Huy and Shi, Guangyao},
  journal={arXiv preprint arXiv:2501.02189},
  volume={1},
  year={2025}
}

@article{
yu2022coca,
title={CoCa: Contrastive Captioners are Image-Text Foundation Models},
author={Jiahui Yu and Zirui Wang and Vijay Vasudevan and Legg Yeung and Mojtaba Seyedhosseini and Yonghui Wu},
journal={Transactions on Machine Learning Research},
issn={2835-8856},
year={2022},
}

@article{jiang2024e5v,
  title={E5-v: Universal embeddings with multimodal large language models},
  author={Jiang, Ting and Song, Minghui and Zhang, Zihan and Huang, Haizhen and Deng, Weiwei and Sun, Feng and Zhang, Qi and Wang, Deqing and Zhuang, Fuzhen},
  journal={arXiv preprint arXiv:2407.12580},
  year={2024}
}

@article{abdin2024phi,
  title={Phi-3 technical report: A highly capable language model locally on your phone},
  author={Abdin, Marah and Aneja, Jyoti and Awadalla, Hany and Awadallah, Ahmed and Awan, Ammar Ahmad and Bach, Nguyen and Bahree, Amit and Bakhtiari, Arash and Bao, Jianmin and Behl, Harkirat and others},
  journal={arXiv preprint arXiv:2404.14219},
  year={2024}
}

@article{alayrac2022flamingo,
  title={Flamingo: a visual language model for few-shot learning},
  author={Alayrac, Jean-Baptiste and Donahue, Jeff and Luc, Pauline and Miech, Antoine and Barr, Iain and Hasson, Yana and Lenc, Karel and Mensch, Arthur and Millican, Katherine and Reynolds, Malcolm and others},
  journal={Advances in neural information processing systems},
  volume={35},
  pages={23716--23736},
  year={2022}
}

@article{achiam2023gpt,
  title={Gpt-4 technical report},
  author={Achiam, Josh and Adler, Steven and Agarwal, Sandhini and Ahmad, Lama and Akkaya, Ilge and Aleman, Florencia Leoni and Almeida, Diogo and Altenschmidt, Janko and Altman, Sam and Anadkat, Shyamal and others},
  journal={arXiv preprint arXiv:2303.08774},
  year={2023}
}

@inproceedings{
hu2021lora,
title={Lo{RA}: Low-Rank Adaptation of Large Language Models},
author={Edward J Hu and yelong shen and Phillip Wallis and Zeyuan Allen-Zhu and Yuanzhi Li and Shean Wang and Lu Wang and Weizhu Chen},
booktitle={International Conference on Learning Representations},
year={2022},
}

@misc{oord2018infonce_contrastive,
    title={Representation Learning with Contrastive Predictive Coding},
    author={Aaron van den Oord and Yazhe Li and Oriol Vinyals},
    year={2018},
    eprint={1807.03748},
    archivePrefix={arXiv},
    primaryClass={cs.LG}
}

@misc{gao2021gradcache,
      title={Scaling Deep Contrastive Learning Batch Size under Memory Limited Setup}, 
      author={Luyu Gao and Yunyi Zhang and Jiawei Han and Jamie Callan},
      year={2021},
      eprint={2101.06983},
      archivePrefix={arXiv},
      primaryClass={cs.LG},
      url={https://arxiv.org/abs/2101.06983}, 
}

@inproceedings{klemmer2025satclip,
  title={Satclip: Global, general-purpose location embeddings with satellite imagery},
  author={Klemmer, Konstantin and Rolf, Esther and Robinson, Caleb and Mackey, Lester and Ru{\ss}wurm, Marc},
  booktitle={Proceedings of the AAAI Conference on Artificial Intelligence},
  volume={39},
  number={4},
  pages={4347--4355},
  year={2025}
}

@article{vivanco2023geoclip,
  title={Geoclip: Clip-inspired alignment between locations and images for effective worldwide geo-localization},
  author={Vivanco Cepeda, Vicente and Nayak, Gaurav Kumar and Shah, Mubarak},
  journal={Advances in Neural Information Processing Systems},
  volume={36},
  pages={8690--8701},
  year={2023}
}

@article{jia2024g3,
  title={G3: an effective and adaptive framework for worldwide geolocalization using large multi-modality models},
  author={Jia, Pengyue and Liu, Yiding and Li, Xiaopeng and Zhao, Xiangyu and Wang, Yuhao and Du, Yantong and Han, Xiao and Wei, Xuetao and Wang, Shuaiqiang and Yin, Dawei},
  journal={Advances in Neural Information Processing Systems},
  volume={37},
  pages={53198--53221},
  year={2024}
}

@article{zhang2024georsclip,
  title={RS5M and GeoRSCLIP: A large scale vision-language dataset and a large vision-language model for remote sensing},
  author={Zhang, Zilun and Zhao, Tiancheng and Guo, Yulong and Yin, Jianwei},
  journal={IEEE Transactions on Geoscience and Remote Sensing},
  year={2024},
  publisher={IEEE}
}

@inproceedings{wang2024skyscript,
  title={Skyscript: A large and semantically diverse vision-language dataset for remote sensing},
  author={Wang, Zhecheng and Prabha, Rajanie and Huang, Tianyuan and Wu, Jiajun and Rajagopal, Ram},
  booktitle={Proceedings of the AAAI Conference on Artificial Intelligence},
  volume={38},
  number={6},
  pages={5805--5813},
  year={2024}
}

@article{liu2024remoteclip,
  title={Remoteclip: A vision language foundation model for remote sensing},
  author={Liu, Fan and Chen, Delong and Guan, Zhangqingyun and Zhou, Xiaocong and Zhu, Jiale and Ye, Qiaolin and Fu, Liyong and Zhou, Jun},
  journal={IEEE Transactions on Geoscience and Remote Sensing},
  year={2024},
  publisher={IEEE}
}

@article{wang2024ringmogpt,
  title={Ringmogpt: A unified remote sensing foundation model for vision, language, and grounded tasks},
  author={Wang, Peijin and Hu, Huiyang and Tong, Boyuan and Zhang, Ziqi and Yao, Fanglong and Feng, Yingchao and Zhu, Zining and Chang, Hao and Diao, Wenhui and Ye, Qixiang and others},
  journal={IEEE Transactions on Geoscience and Remote Sensing},
  year={2024},
  publisher={IEEE}
}

@article{zhou2024geoground,
  title={Geoground: A unified large vision-language model for remote sensing visual grounding},
  author={Zhou, Yue and Lan, Mengcheng and Li, Xiang and Ke, Yiping and Jiang, Xue and Feng, Litong and Zhang, Wayne},
  journal={arXiv preprint arXiv:2411.11904},
  year={2024}
}

@article{zhang2024earthgpt,
  title={Earthgpt: A universal multi-modal large language model for multi-sensor image comprehension in remote sensing domain},
  author={Zhang, Wei and Cai, Miaoxin and Zhang, Tong and Zhuang, Yin and Mao, Xuerui},
  journal={IEEE Transactions on Geoscience and Remote Sensing},
  year={2024},
  publisher={IEEE}
}

@article{li2023rsclip,
  title={RS-CLIP: Zero shot remote sensing scene classification via contrastive vision-language supervision},
  author={Li, Xiang and Wen, Congcong and Hu, Yuan and Zhou, Nan},
  journal={International Journal of Applied Earth Observation and Geoinformation},
  volume={124},
  pages={103497},
  year={2023},
  publisher={Elsevier}
}

@article{mo2023sclip,
  title={S-clip: Semi-supervised vision-language learning using few specialist captions},
  author={Mo, Sangwoo and Kim, Minkyu and Lee, Kyungmin and Shin, Jinwoo},
  journal={Advances in Neural Information Processing Systems},
  volume={36},
  pages={61187--61212},
  year={2023}
}

@inproceedings{pang2025vhm,
  title={Vhm: Versatile and honest vision language model for remote sensing image analysis},
  author={Pang, Chao and Weng, Xingxing and Wu, Jiang and Li, Jiayu and Liu, Yi and Sun, Jiaxing and Li, Weijia and Wang, Shuai and Feng, Litong and Xia, Gui-Song and others},
  booktitle={Proceedings of the AAAI Conference on Artificial Intelligence},
  volume={39},
  number={6},
  pages={6381--6388},
  year={2025}
}

@article{luo2024skysensegpt,
  title={Skysensegpt: A fine-grained instruction tuning dataset and model for remote sensing vision-language understanding},
  author={Luo, Junwei and Pang, Zhen and Zhang, Yongjun and Wang, Tingzhu and Wang, Linlin and Dang, Bo and Lao, Jiangwei and Wang, Jian and Chen, Jingdong and Tan, Yihua and others},
  journal={arXiv preprint arXiv:2406.10100},
  year={2024}
}

@article{hu2025rsgpt,
  title={Rsgpt: A remote sensing vision language model and benchmark},
  author={Hu, Yuan and Yuan, Jianlong and Wen, Congcong and Lu, Xiaonan and Liu, Yu and Li, Xiang},
  journal={ISPRS Journal of Photogrammetry and Remote Sensing},
  volume={224},
  pages={272--286},
  year={2025},
  publisher={Elsevier}
}

@article{zhan2025skyeyegpt,
  title={Skyeyegpt: Unifying remote sensing vision-language tasks via instruction tuning with large language model},
  author={Zhan, Yang and Xiong, Zhitong and Yuan, Yuan},
  journal={ISPRS Journal of Photogrammetry and Remote Sensing},
  volume={221},
  pages={64--77},
  year={2025},
  publisher={Elsevier}
}

@article{bazi2024rsllava,
  title={Rs-llava: A large vision-language model for joint captioning and question answering in remote sensing imagery},
  author={Bazi, Yakoub and Bashmal, Laila and Al Rahhal, Mohamad Mahmoud and Ricci, Riccardo and Melgani, Farid},
  journal={Remote Sensing},
  volume={16},
  number={9},
  pages={1477},
  year={2024},
  publisher={MDPI}
}

@article{Yuan2022RSITMD,
  title={Exploring a Fine-Grained Multiscale Method for Cross-Modal Remote Sensing Image Retrieval},
  author={Yuan, Zhiqiang and Zhang, Wenkai and Fu, Kun and Li, Xuan and Deng, Chubo and Wang, Hongqi and Sun, Xian},
  journal={IEEE Transactions on Geoscience and Remote Sensing},
  volume={60},
  pages={1--19},
  year={2021},
  publisher={IEEE}
}

@article{Lu2017RSICD,
  title={Exploring models and data for remote sensing image caption generation},
  author={Lu, Xiaoqiang and Wang, Binqiang and Zheng, Xiangtao and Li, Xuelong},
  journal={IEEE Transactions on Geoscience and Remote Sensing},
  volume={56},
  number={4},
  pages={2183--2195},
  year={2017},
  publisher={IEEE}
}

@article{Helber2019EuroSAT,
  title={Eurosat: A novel dataset and deep learning benchmark for land use and land cover classification},
  author={Helber, Patrick and Bischke, Benjamin and Dengel, Andreas and Borth, Damian},
  journal={IEEE Journal of Selected Topics in Applied Earth Observations and Remote Sensing},
  volume={12},
  number={7},
  pages={2217--2226},
  year={2019},
  publisher={IEEE}
}

@article{Xia2017AID,
  title={AID: A benchmark data set for performance evaluation of aerial scene classification},
  author={Xia, Gui-Song and Hu, Jingwen and Hu, Fan and Shi, Baoguang and Bai, Xiang and Zhong, Yanfei and Zhang, Liangpei and Lu, Xiaoqiang},
  journal={IEEE Transactions on Geoscience and Remote Sensing},
  volume={55},
  number={7},
  pages={3965--3981},
  year={2017},
  publisher={IEEE}
}

@article{Lobry2020RSVQA,
  title={RSVQA: Visual question answering for remote sensing data},
  author={Lobry, Sylvain and Marcos, Diego and Murray, Jesse and Tuia, Devis},
  journal={IEEE Transactions on Geoscience and Remote Sensing},
  volume={58},
  number={12},
  pages={8555--8566},
  year={2020},
  publisher={IEEE}
}

@inproceedings{Yang2010UCM,
  title={Bag-of-visual-words and spatial extensions for land-use classification},
  author={Yang, Yi and Newsam, Shawn},
  booktitle={Proceedings of the 18th SIGSPATIAL international conference on advances in geographic information systems},
  pages={270--279},
  year={2010}
}

@inproceedings{Xia2018DOTA,
  title={DOTA: A large-scale dataset for object detection in aerial images},
  author={Xia, Gui-Song and Bai, Xiang and Ding, Jian and Zhu, Zhen and Belongie, Serge and Luo, Jiebo and Datcu, Mihai and Pelillo, Marcello and Zhang, Liangpei},
  booktitle={Proceedings of the IEEE conference on computer vision and pattern recognition},
  pages={3974--3983},
  year={2018}
}

@article{Li2020DIOR,
  title={Object detection in optical remote sensing images: A survey and a new benchmark},
  author={Li, Ke and Wan, Gang and Cheng, Gong and Meng, Liqiu and Han, Junwei},
  journal={ISPRS journal of photogrammetry and remote sensing},
  volume={159},
  pages={296--307},
  year={2020},
  publisher={Elsevier}
}

@article{long2021millionaid,
  title={On creating benchmark dataset for aerial image interpretation: Reviews, guidances, and million-aid},
  author={Long, Yang and Xia, Gui-Song and Li, Shengyang and Yang, Wen and Yang, Michael Ying and Zhu, Xiao Xiang and Zhang, Liangpei and Li, Deren},
  journal={IEEE Journal of selected topics in applied earth observations and remote sensing},
  volume={14},
  pages={4205--4230},
  year={2021},
  publisher={IEEE}
}

@article{li2020rsicb,
  title={RSI-CB: A large-scale remote sensing image classification benchmark using crowdsourced data},
  author={Li, Haifeng and Dou, Xin and Tao, Chao and Wu, Zhixiang and Chen, Jie and Peng, Jian and Deng, Min and Zhao, Ling},
  journal={Sensors},
  volume={20},
  number={6},
  pages={1594},
  year={2020},
  publisher={MDPI}
}

@article{zhou2018patternnet,
  title={PatternNet: A benchmark dataset for performance evaluation of remote sensing image retrieval},
  author={Zhou, Weixun and Newsam, Shawn and Li, Congmin and Shao, Zhenfeng},
  journal={ISPRS journal of photogrammetry and remote sensing},
  volume={145},
  pages={197--209},
  year={2018},
  publisher={Elsevier}
}

@inproceedings{qu2016ducmcaption,
  title={Deep semantic understanding of high resolution remote sensing image},
  author={Qu, Bo and Li, Xuelong and Tao, Dacheng and Lu, Xiaoqiang},
  booktitle={2016 International conference on computer, information and telecommunication systems (Cits)},
  pages={1--5},
  year={2016},
  organization={IEEE}
}

@article{Friedman1937ranking,
  title={The Use of Ranks to Avoid the Assumption of Normality Implicit in the Analysis of Variance},
  author={Milton Friedman},
  journal={Journal of the American Statistical Association},
  year={1937},
  volume={32},
  pages={675-701},
  url={https://api.semanticscholar.org/CorpusID:120581754}
}

@article{Friedman1940ranking,
  title={A Comparison of Alternative Tests of Significance for the Problem of \$m\$ Rankings},
  author={Milton Friedman},
  journal={Annals of Mathematical Statistics},
  year={1940},
  volume={11},
  pages={86-92},
  url={https://api.semanticscholar.org/CorpusID:121778036}
}

@inproceedings{wang2022usb,
  title={{USB}: A unified semi-supervised learning benchmark for classification},
  author={Wang, Yidong and Chen, Hao and Fan, Yue and Sun, Wang and Tao, Ran and Hou, Wenxin and Wang, Renjie and Yang, Linyi and Zhou, Zhi and Guo, Lan-Zhe and others},
  booktitle={Advances in Neural Information Processing Systems},
  pages={3938--3961},
  year={2022}
}

@article{loshchilov2017adamw,
  title={Decoupled weight decay regularization},
  author={Loshchilov, Ilya and Hutter, Frank},
  journal={arXiv preprint arXiv:1711.05101},
  year={2017}
}

@misc{xiong2025dofaclip,
      title={DOFA-CLIP: Multimodal Vision-Language Foundation Models for Earth Observation}, 
      author={Zhitong Xiong and Yi Wang and Weikang Yu and Adam J Stewart and Jie Zhao and Nils Lehmann and Thomas Dujardin and Zhenghang Yuan and Pedram Ghamisi and Xiao Xiang Zhu},
      year={2025},
      eprint={2503.06312},
      archivePrefix={arXiv},
      primaryClass={cs.CV},
      url={https://arxiv.org/abs/2503.06312}, 
}

@article{Wei2020HRSID_segmentation,
author = {Wei, Shunjun and Zeng, Xiangfeng and Qu, Qizhe and Wang, Mou and Su, Yung-Yeh and JUN, SHI},
year = {2020},
month = {06},
pages = {1-1},
title = {HRSID: A High-Resolution SAR Images Dataset for Ship Detection and Instance Segmentation},
volume = {8},
journal = {IEEE Access},
doi = {10.1109/ACCESS.2020.3005861}
}

@article{garioud2023flair,
  title={FLAIR: a country-scale land cover semantic segmentation dataset from multi-source optical imagery},
  author={Garioud, Anatol and Gonthier, Nicolas and Landrieu, Loic and De Wit, Apolline and Valette, Marion and Poup{\'e}e, Marc and Giordano, S{\'e}bastien and others},
  journal={Advances in Neural Information Processing Systems},
  volume={36},
  pages={16456--16482},
  year={2023}
}

@InProceedings{SanchezAimar2024ADELLO_LTSSL,
    author="Sanchez Aimar, Emanuel
    and Helgesen, Nathaniel
    and Xu, Yonghao
    and Kuhlmann, Marco
    and Felsberg, Michael",
    title="Flexible Distribution Alignment: Towards Long-Tailed Semi-supervised Learning with Proper Calibration",
    booktitle="Computer Vision -- ECCV 2024",
    year="2025",
    editor="Leonardis, Ale{\v{s}}
    and Ricci, Elisa
    and Roth, Stefan
    and Russakovsky, Olga
    and Sattler, Torsten
    and Varol, G{\"u}l",
    publisher="Springer Nature Switzerland",
    address="Cham",
    pages="307--327",
    isbn="978-3-031-72949-2"
}

@inproceedings{vo2019composing,
  title={Composing text and image for image retrieval-an empirical odyssey},
  author={Vo, Nam and Jiang, Lu and Sun, Chen and Murphy, Kevin and Li, Li-Jia and Fei-Fei, Li and Hays, James},
  booktitle={Proceedings of the IEEE/CVF conference on computer vision and pattern recognition},
  pages={6439--6448},
  year={2019}
}

@inproceedings{psomas2024composed,
      title={Composed Image Retrieval for Remote Sensing}, 
      author={Psomas, B. and Kakogeorgiou, I. and Efthymiadis, N. and Tolias, G. and Chum, O. and Avrithis, Y. and Karantzalos, K.},
      booktitle={IGARSS 2024 - 2024 IEEE International Geoscience and Remote Sensing Symposium}, 
      year={2024}
}

@inproceedings{sun2022visualgrounding,
author = {Sun, Yuxi and Feng, Shanshan and Li, Xutao and Ye, Yunming and Kang, Jian and Huang, Xu},
title = {Visual Grounding in Remote Sensing Images},
year = {2022},
isbn = {9781450392037},
publisher = {Association for Computing Machinery},
address = {New York, NY, USA},
url = {https://doi.org/10.1145/3503161.3548316},
doi = {10.1145/3503161.3548316},
booktitle = {Proceedings of the 30th ACM International Conference on Multimedia},
pages = {404–412},
numpages = {9},
location = {Lisboa, Portugal},
series = {MM '22}
}
